\pdfoutput=1
\documentclass[english,twocolumn,11pt]{IEEEtran}
\usepackage[T1]{fontenc}
\usepackage[latin9]{inputenc}
\usepackage{float}
\usepackage{amsthm}
\usepackage{amsmath}
\usepackage{amssymb}
\usepackage{graphicx}

\makeatletter

\newcommand{\noun}[1]{\textsc{#1}}
\floatstyle{ruled}
\newfloat{algorithm}{tbp}{loa}
\providecommand{\algorithmname}{Algorithm}
\floatname{algorithm}{\protect\algorithmname}

\theoremstyle{plain}
\newtheorem{thm}{\protect\theoremname}
\newenvironment{lyxcode}
{\par\begin{list}{}{
\setlength{\rightmargin}{\leftmargin}
\setlength{\listparindent}{0pt}
\raggedright
\setlength{\itemsep}{0pt}
\setlength{\parsep}{0pt}
\normalfont\ttfamily}%
 \item[]}
{\end{list}}
\theoremstyle{plain}
\newtheorem{lem}[thm]{\protect\lemmaname}

\usepackage{url}

\newcommand{\subparagraph}{}
\usepackage[compact]{titlesec}
\titlespacing*{\section}      
{0pt}{8pt minus 2pt}{6pt minus 2pt}
\titlespacing*{\subsection}
{0pt}{6pt minus 2pt}{4pt minus 2pt}
\titlespacing*{\subsubsection}
{0pt}{0pt}{0pt}
\makeatother

\usepackage{babel}
\providecommand{\lemmaname}{Lemma}
\providecommand{\theoremname}{Theorem}

\begin{document}
\global\long\def\ve#1{\mathbf{#1}}
\global\long\def\mat#1{\mathbf{#1}}
\global\long\def\tr#1{\textrm{tr}\left(#1\right)}

\global\long\def\pd#1#2{\frac{\partial#1}{\partial#2}}

\global\long\def\u{\ve u}
\global\long\def\v{\ve v}
\global\long\def\x{\ve x}
\global\long\def\y{\ve y}

\global\long\def\A{\mat A}

\global\long\def\H{\mat H}
\global\long\def\I{\mat I}
\global\long\def\L{\mat L}
\global\long\def\M{\mat M}
\global\long\def\R{\mat R}
\global\long\def\S{\mat S}
\global\long\def\U{\mat U}
\global\long\def\V{\mat V}
\global\long\def\W{\mat W}
\global\long\def\X{\mat X}
\global\long\def\Y{\mat Y}

\global\long\def\RR{\mathbb{R}}

\global\long\def\lamz{\lambda_{R0min}}
\global\long\def\lamk{\lambda_{Rkmin}}
\global\long\def\lama{\lambda_{Rmin}}
\global\long\def\lamv{\lambda_{Vmin}}
\global\long\def\lamZ{\lambda_{R0max}}
\global\long\def\lamK{\lambda_{Rkmax}}
\global\long\def\lamA{\lambda_{Rmax}}
\global\long\def\lamV{\lambda_{Vmax}}

\global\long\def\nNK{N_{K}}

\global\long\def\NbhdSamples{\mat X_{ni}}
\global\long\def\NbhdSampleW#1{\ve x_{i^{#1}}}
\global\long\def\NbhdSampleB#1{\ve x_{i_{#1}}}

\global\long\def\NbhdWghts{\ve{\omega}_{i}}

\global\long\def\LocalDistMat{\mat Q_{pi}}
\global\long\def\LocalProjMat{\mat W_{pi}}
\global\long\def\LocalLapMat{\mat L_{i}}

\global\long\def\LocalSeleMat{\mat S_{i}}

\global\long\def\MahaDist#1#2#3{d_{#1}(#2,#3)}

\global\long\def\SubSampleNum{N_{K}}
\global\long\def\SubProjMatk{\W_{k}}
\global\long\def\SubSamLapk{\R_{k}}
\global\long\def\SubLapMatk{\mat L_{k}}

\global\long\def\IdealProjMat{\W_{0}}
\global\long\def\EmpProjMat{\W_{G}}
\global\long\def\SubEmpProjMat{\W_{N}}
\global\long\def\AggSubEmpProjMat{\hat{\mat W}}
\global\long\def\OrthAggProjMat{\tilde{\mat W}}

\global\long\def\fnObj{\hat{F}}
\global\long\def\fnObjA{\fnObj'}
\global\long\def\fnDeObj{\hat{M}}
\global\long\def\fnObjk{F_{k}}
\global\long\def\fnDeObjk{M_{k}}

\global\long\def\MaxEig{\lambda_{max}}
\global\long\def\MinEig{\lambda_{min}}

\global\long\def\MinEigRk{\gamma_{k}}
\global\long\def\MinEigAllk{\gamma_{T}}
\global\long\def\MinEigR{\gamma_{R}}
\global\long\def\MinEigV{\gamma_{V}}

\global\long\def\MaxEigRk{\Gamma_{k}}
\global\long\def\MaxEigAllk{\Gamma_{T}}
\global\long\def\MaxEigR{\Gamma_{R}}
\global\long\def\MaxEigV{\Gamma_{V}}

\title{A Distributed Approach towards Discriminative Distance Metric Learning}

\author{Jun Li, Xun Lin, Xiaoguang Rui, Yong Rui \emph{Fellow, IEEE} and Dacheng Tao \emph{Senior Member, IEEE}%
\thanks{This work was supported by the Australian Research Council
under Projects FT-130101457, DP-140102164, and LP-140100569.

J. Li and D. Tao are with the Centre for Quantum Computation and Intelligent
Systems, Faculty of Engineering and Information Technology, University of
Technology, Sydney, Ultimo, NSW 2007, Australia (e-mail: \{jun.li, dacheng.tao\}@uts.edu.au). 

X. Lin is with Shenzhen Institute of Advanced Technology, Chinese Academy of
Sciences, Shenzhen, China (e-mail: xlin.scut@gmail.com).

X. Rui is with China Research Lab, IBM, Beijing, China (e-mail: ruixg@cn.ibm.com).

Y. Rui is with Microsoft Research Asia, Beijing, China (e-mail: yongrui@microsoft.com)

\textcopyright 20XX IEEE. Personal use of this material is permitted.
Permission from IEEE must be obtained for all other uses, in any current or
future media, including reprinting/republishing this material for advertising
or promotional purposes, creating new collective works, for resale or
redistribution to servers or lists, or reuse of any copyrighted component of
this work in other works.
}}
\maketitle
\begin{abstract}
Distance metric learning is successful in discovering intrinsic relations
in data. However, most algorithms are computationally demanding when
the problem size becomes large. In this paper, we propose a discriminative
metric learning algorithm, and develop a distributed scheme learning
metrics on moderate-sized subsets of data, and aggregating the results
into a global solution. The technique leverages the power of parallel
computation. The algorithm of the aggregated distance metric learning
(ADML) scales well with the data size and can be controlled by the
partition. We theoretically analyse and provide bounds for the error
induced by the distributed treatment. We have conducted experimental
evaluation of ADML, both on specially designed tests and on practical
image annotation tasks. Those tests have shown that ADML achieves
the state-of-the-art performance at only a fraction of the cost incurred
by most existing methods.

\noun{Index Terms} -- parallel computing, distance metric learning,
online learning 
\end{abstract}

\section{Introduction}

Comparing objects of interest is a ubiquitous activity and defining
characteristic of learning-based systems. The comparison can be explicit,
as in the nearest neighbour rule, or be encoded in a learned model,
as in the neural networks. In all cases, to let the past experience
have any help in making decisions about unseen objects, one must compare
the objects to those with known information. A natural measure is
the Euclidean distance. Despite its wide application, the Euclidean
distance may not suit all problems. For example, Euclidean distance
is directly affected by the scaling of individual features in the
representation of the data. Features of high multitude have strong
influence on the measure of similarity regardless its relevance to
the task. Without accounting for relevance, Euclidean distance is
particularly problematic when the data is of high dimension, and the
informative structures of the data population are difficult to distinguish
from meaningless fluctuations.

The problem of the Euclidean distance suggests to adapt the metric
in the learning process, so that distance measure is conducive to
the subsequent recognition tasks. Along this line of research, a family
of distance metric learning techniques has been developed \cite{Xin03}\cite{She02}\cite{Hoi06}\cite{Gol05}\cite{Wei06}\cite{Fouad13}\cite{Shen10b},
and proven useful in a number of practical applications \cite{Hoi00}\cite{Kan06}.
However, a major difficulty of metric learning arises from the time
and space cost of those algorithms. To construct an effective distance
metric, relevance of the raw features should be evaluated w.r.t. each
other, instead of individually. Given $d$ raw features, this means
$d^{2}$ covariance terms to be dealt with. In cases where $d$ is
large but the number of samples, $N$, is moderate, an alternative
formulation of the problem allows learning metric from pair-wise correlations
between samples, which entails a complexity of $N^{2}$. However,
the problem becomes irreducibly complex when both $d$ and $N$ are
large, which is common in practical problems. In fact, due to technical
necessities such as iterations in optimisation, a realistic metric
learning algorithm generally involves a complexity of the cubic order,
such as $d^{3}$ or $dN^{2}$, rather than squared one, which further
limits the scalability of metric learning.

The focus of this paper is to address the seemingly inexorable complexity
of distance metric learning. We first develop a discriminative metric
learning method, where categorical information are utilised to direct
the construction of the distance metric. More importantly, the metric
learning algorithm embraces the \textquotedblleft divide-and-conquer\textquotedblright{}
strategy to deal with large volume of high-dimensional data. We derive
a distributed and parallel metric learning scheme, which can be implemented
in consistence with the MapReduce \cite{mapredu} computational framework.
In particular, we split a large sample into multiple smaller subsets.
The basic discriminative metric learning is applied on each of the
subsets. The separately learned metrics are summarised into the final
result via an aggregation procedure, and the scheme is named aggregated
distance metric learning (ADML), accordingly.

Corresponding to the Map steps in the MapReduce paradigm, the computation
on the subsets are independent with each other, and can be conducted
in parallel tasks. The data can be stored in a distributed system,
and no individual subroutine needs access to the whole data at once.
So the technique is less affected by the physical storage in dealing
with large volume of data. The aggregation step represents the Reduce
step in MapReduce, where the input of the aggregation algorithm is
the learned metrics on the subsets. Aggregation takes sum of those
learned metrics, where the operation collapses the inputs arriving
in arbitrary order, taking only moderate space given the subspace
representation of the metrics.

The whole scheme scales well with the dimension and sample size of
the data. The basic discriminative metric learning algorithm adopts
subspace representation of the learned metric. If the sizes of individual
subsets $N_{K}$ and the subspace dimension $q$ are fixed, the space
and time taken by the metric learning on the subsets grow linearly
with respect to the input dimension $d$. The aggregation time grows
linearly with $d$ as well. Thus in theory, the ideal implementation
of ADML has desirably benign time and space cost. The time cost is
linear with respect to the input dimension $d$ and \emph{independent}
to the total sample size $N$%
\footnote{The first step of aggregation sums up the metrics learned on subsets.
The summation could be parallelised in theory, which renders the cost
of the entire scheme independent to the sample size $N$. However,
in practice, the parallel tasks are limited by the physical computing
power and communication, where the time cost of summation is negligible
compared to the rest of the learning. Thus practical implementation
of the aggregation uses serial summation, whose time cost is linear
w.r.t. $N$.%
}, and the space cost is the size of the data $d\times N$ in terms
of storage and $\max(d\times q,N_{K}^{2})$ in terms of the volatile
memory for on-the-fly computation, where $d\times q$ represents the
size of the learned metric, and $d$ and $N$ are large compared to
$q$ and $N_{K}$. For practical computational facilities, where parallel
tasks are limited and incurs communication overheads, the learning
and aggregation of metrics on the subsets can only be partially parallelised.
Thus the time cost by ADML grows linearly with, instead of being independent
to, the sample size $N$. This practically achievable time complexity
still scales better with the problem size compared with wholistic
methods. 

We provide theoretical guarantee for the distributed computation scheme.
In particular, the upper bound has been provided for deviation by
which the metric obtained by ADML can differ from the one would be
resulted if the discriminative metric learning were applied on the
whole data. The bound of deviation supports the usage of distributed
computation in metric learning, and caps the price we have to pay
for the gain of efficiency. 

The effectiveness of ADML has also been demonstrated by empirical
study. We test the method using both synthetic data and on a practical
image annotation task. The empirical results corroborate our motivation
for ADML, that the distributed learning achieves much superior efficiency
with little or no sacrifice of accuracy compared to state-of-the-art
metric learning methods.

\section{Background\label{sec:rev}}

The need for appropriate similarity measure is rooted in the fundamental
statistical learning theory, in particular in the interplay between
complexity and generalisation. In general, a learning algorithm produces
a predictor by matching a set of hypotheses against observed data.
The theory tells that the confidence about the predictor\textquoteright s
performance on unseen data is affected by the complexity of the hypothesis
set \cite{Vap98}. While the complexity of a hypothesis (function)
set is gauged by the possible variations on the domain of input. Therefore,
the way of measuring similarity between the data samples essentially
characterises the hypothesis set and affects how a learned predictor
generalises to unseen data.

It has been noticed that the intuitive measurement of Euclidean metric
becomes unreliable for high dimensional data, i.e. the number of raw
features is large \cite{Liu06}. Therefore, recently much attention
has been paid to adapting the distance metrics to the data and analysis
tasks. In \cite{Xin03}, Xing et al. proposed to adaptively construct
a distance measure from the data, and formulated the task as a convex
optimisation problem. The technique was successfully applied to clustering
tasks. An effective way of learning a metric and simultaneously discovering
parsimonious representation of data is to construct a mapping, which
projects the data into a subspace in which the Euclidean distance
reproduces the desired metric. The link between metric learning and
data representation connects the proposed method to a broad context.
In \cite{Yeung07}, metric learning is formulated as a kernel learning
problem, and solved via convex optimisation. Relevance component analysis
(RCA) \cite{She02} learns a transformation with accounting for the
equivalent constraints presenting in the data, i.e. pairs of samples
are known to be of the same category and thus preferably to be measured
as close to each other by the objective metric. In \cite{Hoi06},
Hoi et al. proposed discriminant component analysis (DCA), which equips
RCA with extra capability of dealing with negative constraints, sample
pairs to be scattered far apart by a desirable metric. Both RCA and
DCA can be solved by eigenvalue decomposition, which is practically
faster than the convex optimisation in \cite{Xin03}. When the ultimate
goal of learning a distance metric is to preserve or enhance discriminative
information, research has shown that the local geometry in the neighbourhoods
of individual samples is effective. The raw features can be assessed
by \cite{Dom02,Has96} by their contribution to the posterior probabilities
at individual samples. In \cite{Gol05}, a learning scheme has been
proposed to optimise classification based on nearest neighbour rule.
Large margin objective has also been proven effective in deriving
supervised metric learning approaches \cite{Wei06,Shen10b}. In \cite{Fouad13},
a scheme has been developed to incorporate auxiliary knowledge to
assist metric learning. More comprehensive overview of related research
can be found in \cite{Liu06}. As we have discussed in the last section,
a major concern of most existing metric learning approaches is the
scalability. For example, the constrained convex programming \cite{Xin03,She02,Hoi06,Gol05,Wei06}
limits their applicability in large-scale practical problems. 

Online processing is another strategy of solving large scale problem
\cite{Grippo00}, which is complement to parallel computing and often
of practical interest when dealing with data from the web. Online
distance metric learning algorithms has been proposed \cite{Dav07}\cite{Sha04}.
These online metric learning techniques are based on serial steps,
where the metric is updated after new observations arrive. Serial
online schemes are able to deal with large problems, but slow to arrive
the final solution. To improve efficiency, parallel implementation
is a natural strategy. Many basic operations have fast parallelised
implementation, such as Intel's Math Kernel Library; and some learning
techniques are parallelised at the algorithm level \cite{Cat08}\cite{Rai09}\cite{Chu07}.
While ADML is a novel metric learning method supporting fully distributed
realisation, where subset metrics can be computed on different subsets
simultaneously, and with reduced computational cost.

\section{Discriminative distance metric learning \label{sec:DDML}}

We are concerned with constructing a Mahalanobis distance metric in
the data feature space, which incorporates the discriminative information,
i.e. the class membership of individual samples. In particular, the
Mahalanobis distance between $\x_{i},\x_{j}\in\mathbb{R}^{d}$ is
defined by a positive semi-definite matrix $\mathbf{Q}\in\mathbb{R}^{d\times d}$,
$d_{\mathbf{Q}}^{2}(\x_{i},\x_{j})=(\x_{i}-\x_{j})^{T}\mathbf{Q}(\x_{i}-\x_{j}).$
The goal of the proposed discriminative distance metric learning (DDML)
is to find a matrix $\mat Q$, so that $d_{\mathbf{Q}}$ between samples
of the same class tends to be small and that between samples of different
classes tends to be large. Such a distance metric will benefit subsequent
tasks such as classification or recognition. 

However, a universally satisfactory Mahalanobis distance metric $\mat Q$
is usually difficult to obtain. Practical data are often complex objects
consisting of many features, where the number of features, $d$, is
much greater than the actual degrees of freedom. Thus the variations
in the data span nonlinear low-dimensional manifolds precluding global
Mahalanobis distance metric. To preserve the nonlinear structures
in the data, we set up the learning objective locally and let a local
metric encodes both the discriminative information and the local geometric
structure of the data. The overall distance metric is computed by
unifying all the local metrics in an optimisation framework.

In particular, the cost of a distance metric is defined on data \emph{patches}
associated with individual samples. Given a sample $\x_{i}$, the
corresponding patch consists of a small number of samples that are
close to $\x_{i}$ and carrying discriminative information. In particular,
we select the $k_{W}$ nearest neighbours $\x_{N_{i}^{W}}$ from the
same class as $\x_{i}$, and select $k_{B}$ nearest neighbours $\x_{N_{i}^{B}}$
from different classes. The patch, $\X_{N_{i}}$, is the joint of
$\{\x_{i}\}\cup\x_{N_{i}^{W}}\cup\x_{N_{i}^{B}}$.  

The local criterion of metric learning on $\X_{N_{i}}$ is motivated
by encoding the discriminative information into the geometry induced
by the objective metric. A local metric $\mat Q_{i}$ specifies the
within- and between-class distances on $\X_{N_{i}}$ as
\begin{align}
D^{\{W,B\}}(\X_{N_{i}},\mat Q_{i})= & \sum_{j\in N_{i}^{\{W,B\}}}d_{\mat Q_{i}}^{2}(\x_{i},\x_{j})
\end{align}
Furthermore, to account for the locally linear geometry of the data
manifold, $\mat Q_{i}$ is considered as being induced by a subspace
projection $\W_{i}$, where $d_{\mat Q_{i}}^{2}(\x_{i},\x_{j})=\|\W_{i}^{T}(\x_{i}-\x_{j})\|^{2}$.
Note that such a subspace representation of the Mahalanobis distance
implies a concise parameterisation of $\mat Q_{i}=\mat W_{i}\mat{W_{i}}^{T}$.

Therefore, the principle of metric learning becomes an optimisation
over local transformation $\W_{i}$,

\begin{align}
\W_{i} & =\arg\min_{\W}\big\{{\textstyle \sum_{j\in N_{i}^{W}}}\|\W^{T}(\x_{i}-\x_{j})\|^{2}\label{eq:localProjObj}\\
 & \qquad\ \ \ -\beta{\textstyle \sum_{j\in N_{i}^{B}}}\|\W^{T}(\x_{i}-\x_{j})\|^{2}\big\}\nonumber \\
 & =\arg\min_{\W}\tr{\W^{T}\X_{N_{i}}\L_{i}\X_{N_{i}}^{T}\W}\nonumber 
\end{align}
where $\beta$ is a trading-off between small $D^{W}$ and large $D^{B}$;
and $\L_{i}$ is obtained by organising the coefficients of the quadratic
form, 
\begin{eqnarray*}
 & \LocalLapMat= & {\textstyle \left[\begin{array}{cc}
{\displaystyle {\textstyle \sum_{j}}\left[\NbhdWghts\right]_{j}} & -\NbhdWghts^{T}\\
-\NbhdWghts & \mathrm{diag}(\NbhdWghts)
\end{array}\right]}
\end{eqnarray*}
where $\omega_{i}$ is a row vector concatenated by $N_{i}^{W}$ $[1]$'s
and $N_{i}^{B}$ $[-\beta]$'s, and $\mathrm{diag}(\cdot)$ represents
an operator generating an $n\times n$ diagonal matrix from an $n$-dimensional
vector.

Unifying the optimisations on individual local patches, we reach the
objective function for global distance metric learning,
\begin{align}
\W_{G} & =\arg\min_{\W}\sum_{i}\tr{\W^{T}\X_{N_{i}}\L_{i}\X_{N_{i}}^{T}\W}\nonumber \\
 & =\arg\min_{\W}\,\tr{\W^{T}\X\L\X^{T}\W},\label{eq:gobj}
\end{align}
where $\L=\sum_{i}\LocalSeleMat\LocalLapMat\LocalSeleMat^{T}$ and
$\LocalSeleMat$ is an $|\X_{N_{i}}|\times N$ matrix selecting the
samples in a local patch from the data set, so that $\X_{N_{i}}=\X\LocalSeleMat$.

The DDML objective (\ref{eq:gobj}) is readily solved by eigenvalue
decomposition, where the columns of $\W$ are of the vectors associated
with the smallest eigenvalues of $\X\L\X^{T}$. Generally, the eigenvalue
decomposition involved in solving $\W$ in (\ref{eq:gobj}) will be
more efficient than methods based on iterative convex optimisation
methods, e.g. NCA \cite{She02} and LMNN \cite{Wei06}. 

For most practical applications, DDML is still facing several difficulties,
because both the volume and dimension of practical data are usually
high, and also because in real-life applications, the data is often
organised in distributed databases and acquired as streams.  In the
following section, we will address these problems by applying the
idea of ``divide and conquer'' \cite{Lin11} to DDML and propose
a scalable aggregated distance metric learning algorithm.

\section{Aggregated Distance Metric Learning \label{sec:ADML}}

In this section, we first introduce the distributed computation of
DDML, aggregated distance metric learning (ADML), and then prove that
metric learned by ADML is consistent with the result of performing
DDML directly on the entire dataset. We further improve the efficiency
of ADML by introducing an alternative aggregation technique, and show
its consistency attributes. The time and spatial complexities are
discussed for the proposed algorithms with an important observation
that we can compute discriminative metric learning independent of
the data size and only linearly depending on the data dimension.

\subsection{Aggregated Distance Metric Learning}

The idea of ADML can be sketched as follows: dividing the whole data
into several subsets, performing DDML on each of the subset and finally
aggregating the results into a consolidate solution. It is important
not to confuse the split of data in ADML with the use of local patch
in DDML. We recall that the objective function in DDML is defined
on a patch-by-patch base to account for the local geometry of the
data manifold, where each patch is constructed with regard to the
entire dataset. While in ADML, subsets are generally \emph{not} associated
with local geometry (preferably so), and only one of the subsets is
visible to a DDML algorithm.

We randomly split a dataset $\X$ of $N$ samples into $K$ subsets
$\X_{k=1,\dots,K}$.  This reduces the cost of constructing patches
in metric learning: in ADML, the patch of a sample is constructed
by finding nearest neighbours within the subset to which the sample
is belonging. 

Let the objective function of learning $\W$ in (\ref{eq:gobj}) be
$F(\W)$. Using the similar form as $F$, we can derive a local objective
function with respect to a subset $\X_{k}$,
\begin{align}
F_{k}(\W) & =\tr{\W^{T}\SubSamLapk\W},\label{eq:W.obj}
\end{align}
where $\SubSamLapk=\X_{k}\L_{k}\X_{k}^{T}$ and $\L_{k}$ is defined
similarly to $\L$ in (\ref{eq:gobj}), but is confined within $\X_{k}$.
The metric on subset $k$ is then characterised by $F_{k}$, where
we denote the local solution be $\W_{k}$.

The next task is to consolidate the local solutions $\{\W_{k}\}$
into a global metric $\W_{A}$. Straightforward linear combination
would be the most intuitive approach. However, $\{\W_{k}\}$ are solutions
to locally defined optimisation problems and linear interpolation
may damage their optimality and yield invalid solutions. Therefore,
we design the aggregation based on the optimal conditions of the objective
functions. A local solution $\W_{k}$ is a stationary point of the
corresponding $F_{k}$, where the gradient vanishes,
\begin{align}
 & \pd{F_{k}(\W_{k})}{\W_{k}}=2\SubSamLapk\W_{k}=0,\ \textrm{for}\ k=1,\dots,K.\label{eq:stat-point-cond}
\end{align}
Recall that $\R_{k}$ is defined in (\ref{eq:W.obj}). For a global
solution $\W_{G}$, it is ideal for it to fulfil condition (\ref{eq:stat-point-cond})
in all local subsets $k=1,\dots,K$, which is generally impossible.
Thus secondarily, we want the violations to cancel out among all the
$K$ subsets and the sum of local gradients to vanish
\begin{equation}
\sum\limits _{k=1}^{K}\R_{k}\W_{A}=0.\label{eq:TotalProjMatGrad}
\end{equation}
Directly summing up (\ref{eq:stat-point-cond}) over $k$, we reach
\begin{equation}
\sum\limits _{k=1}^{K}\R_{k}\W_{k}=0.\label{eq:LocalProjMatGrad}
\end{equation}
Comparing (\ref{eq:LocalProjMatGrad}) and (\ref{eq:TotalProjMatGrad})
gives an aggregation rule: the overall solution $\W_{A}$ satisfies

\begin{align}
0 & =\sum_{k=1}^{K}\R_{k}\W_{k}-\sum_{k=1}^{K}\R_{k}\W_{A}\label{eq:StatCondForTotalProjMat}\\
\textrm{thus }\W_{A} & =\left(\sum_{k=1}^{K}\R_{k}\right)^{-1}\left(\sum_{k=1}^{K}\SubSamLapk\SubProjMatk\right).\label{eq:SolTotalProjMat}
\end{align}
The computation of (\ref{eq:SolTotalProjMat}) is intuitively plausible:
the aggregation has the form of interpolation of the local $\W_{k}$
with the ``weights'' being $\R_{k}$. We will show that the aggregation
also enjoys desirable theoretical attributes in following discussions. 

There are two further remarks regarding the derivation and computation
of the aggregation. First, directly solving for $\SubProjMatk$ following
(\ref{eq:W.obj}) involves eigenvalue decomposition of a $d\times d$
matrix, which is expensive when $d$ is large. The cost can be alleviated
by exploiting that $\R_{k}$ is known to be decomposed as $\X_{k}\L_{k}\X_{k}^{T}$.
In the scenario of a large $d$, a subset $\X_{k}$ generally has
$|\X_{k}|<d$ and samples independent vectors in $\RR^{d}$. Thus
$\W_{k}$ can be represented as $\X_{k}\U_{k}$. Solving for $\W_{k}$
becomes finding $\U_{k}$ such that 
\begin{align*}
\X_{k}\L_{k}\X_{k}^{T}\X_{k}\U_{k} & =\X_{k}\U_{k}\mat{\Lambda}_{k},
\end{align*}
where $\mat{\Lambda}_{k}$ is a diagonal matrix. It is sufficient
to find a $\U_{k}$ where $\L_{k}\X_{k}^{T}\X_{k}\U_{k}=\U_{k}\mat{\Lambda}_{k}$,
where translates to an eigenvalue decomposition of an $|\X_{k}|\times|\X_{k}|$
matrix $\L_{k}\X_{k}^{T}\X_{k}$.  

The second remark is on the relation between the two characteristics
$\W$, the solution of local DDML: (i) $\W$ is obtained by eigenvalue
decomposition and minimises (\ref{eq:W.obj}) and (\ref{eq:gobj})
and (ii) $\W$ satisfies stationary point condition (\ref{eq:stat-point-cond}).
We do not include the orthogonal constraint on $\W$ in (\ref{eq:stat-point-cond}),
because at the aggregation stage, we have already got the local solutions
complying with the constraint. In fact, if we consider one column
of $\W$ and formulate the orthogonal constraint using Lagrange multiplier,
$L=\ve w^{T}\R\ve w-(\lambda\ve w^{T}\ve w-1)$, then the stationary
condition with respect to $L$ leads to the solution of eigenvalue
decomposition.

\subsection{Consistency of Aggregation}

In this section, we prove that the aggregated distance metric is consistent
with the one we would achieve if we performed the learning using a
wholistic approach. The essential idea in brief is as follows. 

The law of large numbers dictates that the empirical solution of (\ref{eq:gobj}),
say, $\W_{G}$, approaches to an ``ideal solution'' $\W_{0}$, the
solution to (\ref{eq:gobj}) if we could have constructed the problem
using the entire data population as $\X$. We first notice that the
same argument applies to the local solutions $\{\W_{k}\}$ -- solution
of (\ref{eq:gobj}) on a subset $\X_{k}$ approaches $\W_{0}$ as
$\X_{k}$ start including infinitely many random samples, which is
the case because of the random split of data. Thus to show the aggregated
solution $\W_{A}$ approaches the wholistic solution $\W_{G}$, we
can show that both of them approach $\W_{0}$. We prove $\W_{A}$
approaches $\W_{0}$ through $\{\W_{k}\}$. In fact, our key contribution
is to establish that given some target $\hat{\W}$, the distance between
the aggregated solution $\W_{A}$ and $\hat{\W}$ is bounded in terms
of the distance between the local solutions $\W_{k}$ and $\hat{\W}$
(by a constant factor). The result is formally specified by the following
theorem.

\begin{thm}
\label{the:3.1} Let $\{\W_{k}\}$ be the local solutions to (\ref{eq:W.obj})
on the subsets. If $\R=\sum_{k}\R_{k}$ in (\ref{eq:SolTotalProjMat})
is invertible and $\W_{G}$ is the aggregation of $\{\W_{k}\}$, we
have $\|\W_{A}-\hat{\W}\|\leq S_{1}\sup_{k}\{\|\W_{k}-\hat{\W}\|\}$
for some target $\hat{\W}$, where\end{thm}
\begin{itemize}
\item $\|\cdot\|$ \emph{represents the spectral norm of a matrix, $\|\mat A\|=\sqrt{\lambda_{max}(\mat A^{T}\mat A)}$,
where $\lambda_{max/min}$ stands for the maximum or minimum eigenvalue
of a matrix; }
\item \emph{$S_{1}=\frac{K\max_{k}\{\lambda_{max}(\R_{k})\}}{\lambda_{min}(\R)}$,
and $K$ is the number of subsets.}
\end{itemize}
\emph{Moreover, if each $\R_{k}$ is positive definite, we have $\|\W_{A}-\hat{\W}\|\leq S_{2}\max_{k}\{\|\W_{k}-\hat{\W}\|\}$,
where $S_{2}=\frac{\max_{k}\{\lambda_{max}(\R_{k})\}}{\min_{k}\{\lambda_{min}(\R_{k})\}}$}.

We sketch the proof of the theorem as follows, and provide technical
details in Appendix. The first part, i.e. $\|\W_{A}-\hat{\W}\|\leq S_{1}\max_{k}\{\|\W_{k}-\hat{\W}\|\}$,
is achieved by applying matrix triangle inequality \cite{Hor90} and
then manipulating the max/min eigenvalues for each item in the summation.
For the second part, the proof is similar to the first part, additionally,
with a bound of $\lambda_{min}$ of $\R$ being derived in terms of
$\lambda_{min}$ of individual $\R_{k}$.

By replacing the target matrix $\hat{\W}$ with $\W_{0}$ and $\W_{G}$,
Theorem \ref{the:3.1} straightforwardly shows the consistency of
ADML.

\subsection{\label{sub:ADML2}Orthogonalised Aggregated DML}

We have introduced ADML algorithm and shown that its divide-and-conquer
scheme is valid in theory. However, for practical implementation,
inverting $\R=\sum_{k}\SubSamLapk$ for aggregation is a computationally
demanding step if the dimension of data is high. In this section,
we will introduce an efficient operation to replace the matrix inversion,
as well as show that the alternate implementation of aggregation enjoys
consistency attributes similar to what we have established above.

We recall our earlier discussion that $\left(\sum_{k}\SubSamLapk\right)^{-1}$
in (\ref{eq:SolTotalProjMat}) serves as a normalising factor in a
matrix interpolation. To avoid computing the inversion, we employ
singular value decomposition (SVD) for the mixing of the mapping matrices
in the weighted sum $\sum_{k}\SubSamLapk\SubProjMatk$. SVD is relatively
inexpensive compared to inversion, because in general we let $\W$
be a mapping to a subspace of much lower dimension than the raw data
space, and $\W$ has much less columns than rows. Formally, we define
the new aggregation rule by the following SVD
\begin{align}
\W_{A}\mat D\V^{T} & =\sum_{k=1}^{K}\SubSamLapk\SubProjMatk,\label{eq:agg-by-svd}
\end{align}
where $\mat D$ is a diagonal matrix and $\V$ has orthonormal columns.
We summarise the steps for a practical implementation of ADML in Algorithm
\ref{alg:adml}. The consistency attributes are established by the
following theorem.
\begin{thm}
\label{the:alter-agg-consis}Given locally learned $\W_{k=1,\dots,K}$
and aggregated $\W_{A}$ by rule in (\ref{eq:agg-by-svd}), we have
$\|\W_{A}-\hat{\W}\|\leq KS_{3}\max_{k}\left\{ \left\Vert \W_{k}-\hat{\W}\right\Vert \right\} +S_{3}+1$
for some target $\hat{\W}$, where $S_{3}=\lambda_{max}(\R)/\min\{\mathrm{diag}(\mat D)\}$. 
\end{thm}
A proof of Theorem \ref{the:alter-agg-consis} can be reached following
the similar route of proving Theorem \ref{the:3.1}. The details are
provided in Appendix. The consistency of the alternated aggregation
can then be established.

\subsection{\label{sub:Comparison}Temporal and Spatial Complexity}

\begin{table}
\protect\caption{Step-wise Time Complexity\label{tbl:1}}

\begin{centering}
\includegraphics[width=6.6cm]{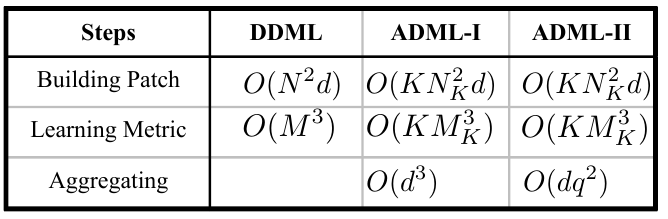}
\par\end{centering}

\noindent {\footnotesize{}ADML-I represents the aggregation by the
rule of (\ref{eq:SolTotalProjMat}); and ADML-II represents aggregation
by rule (\ref{eq:agg-by-svd}). $M=\min\{N,d\},\ M_{K}=\min\{N_{K},d\}$,
$N$: total number of samples, $N_{K}$: number of samples in individual
subsets $d$: data dimension, $K$: number of subsets, $q$: number
of columns in $\W$.}
\end{table}

One of the main motivations of ADML is to scale metric learning up
for practical applications. Table \ref{tbl:1} lists the time complexity
of the steps of the learning algorithms discussed in this section.
Note that the complexity of ADML algorithms are for serial processing
of $K$ subsets, each of size $N_{K}$. However, ADML can be parallelised.
Independent sub-routines of subset metric learning can be carried
out simultaneously. The time of the metric learning on subsets can
be accelerated by a factor up to $K$. The spatial complexity of the
aggregated algorithms is also smaller than that of the straightforward
implementation of DDML. In a parallel implementation, each computational
node needs only to store one subset in its memory, and the computation
takes less memory as well. 

\section{Experiments\label{sec:Experiments}}

In this section, we conduct experimental studies on synthetic and
real data to validate the proposed metric learning technique. In all
the tests, the aggregation rule discussed in Sec. \ref{sub:ADML2}
is employed (ADML-II, referred to as ADML hereafter), because it has
superior temporal and spatial efficiency. We assess the methods by
how the learned metric aids data representation, as well as their
empirical costs of computational resources.

\subsection{Experiments on synthetic data\label{sub:toy}}

\begin{figure}
\begin{centering}
\includegraphics[width=0.3\textwidth]{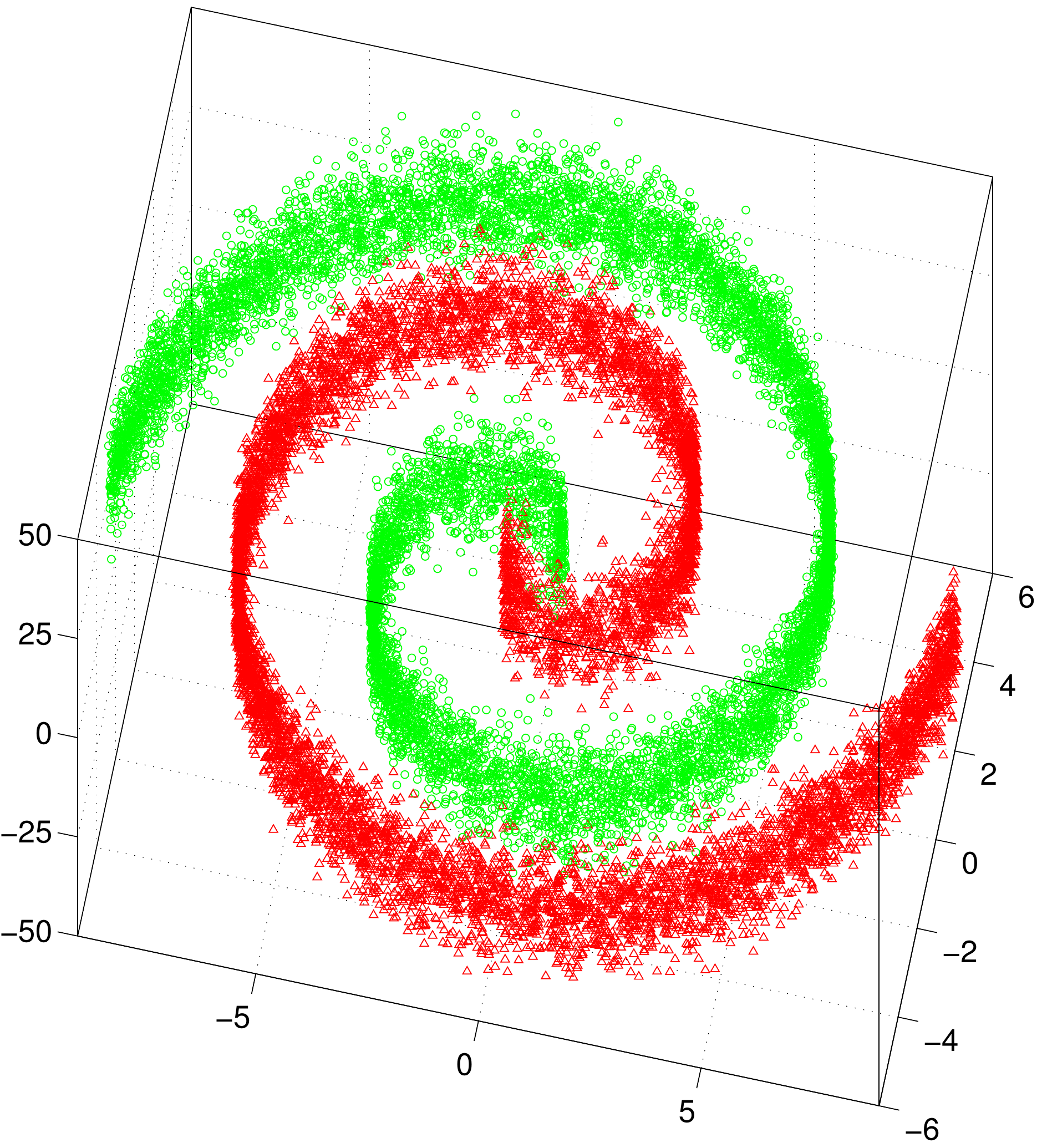}
\par\end{centering}

\protect\caption{A set of data points drawn from the synthetic distribution. \label{fig:syndata}
The green and red points belong to two classes, being slightly disturbed
from two coiled surfaces. Note the range of the data distribution
along the Z-axis (the one that is mostly perpendicular to the canvas)
is greater than those along the X-Y axis. }

\end{figure}

\begin{figure}
\begin{centering}
\includegraphics[height=1.5in]{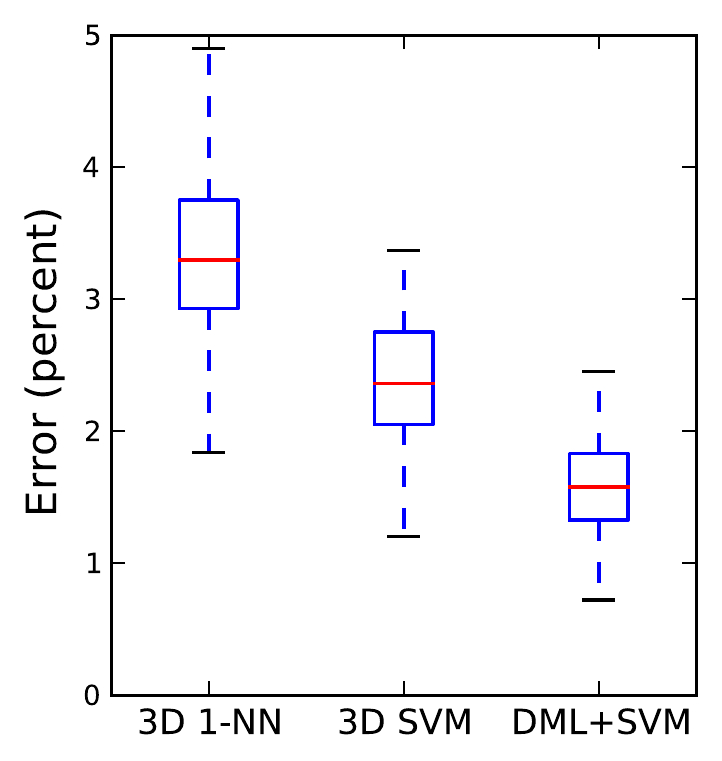}\includegraphics[height=1.5in]{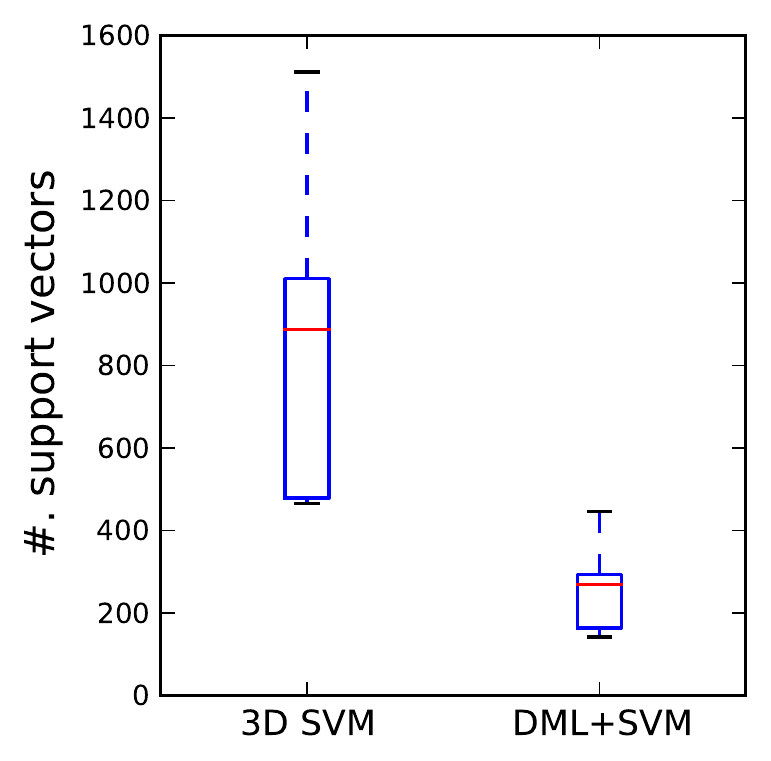}
\par\end{centering}

\begin{centering}
{\footnotesize{}(a)\hspace{1.5in}(b)}
\par\end{centering}{\footnotesize \par}

\protect\caption{How distance metric learning and subspace mapping affect classification.
\textbf{\label{fig:dmlsvm}(a)} comparison between three classification
tests: nearest neighbour rule and SVM in the raw 3D data space, and
SVM in projected 2D space. \textbf{(b)} number of support vectors
the SVMs used in 3D and 2D spaces.}

\end{figure}

The algorithm is first tested on synthetic data consisting of two
classes of 3D points distributed on two curved surfaces. The surfaces
are coiled together in the 3D space, and the data points deviate away
from the surfaces by small disturbances. The experiment has been repeated
20 times using randomly sampled datasets, each containing approximately
17,000 samples. Fig. \ref{fig:syndata} illustrates one set of samples
drawn from the population. In this experiment, we will first verify
the principal motivation that an appropriate distance metric helps
analysis, before turning to the distributed implementation of ADML. 

The first test is by using the basic method developed in Sec. \ref{sec:DDML},
which allows us to examine the effect of DML without complication
caused by any divide-aggregation scheme. The metric is represented
by a $3\times2$ matrix $\W$, because the variance relevant to classification
spans across two dimensions (the X- and Y-axis). Although DML naturally
lends itself to classification by nearest neighbour rule, which will
be our classifier of choice in the following experiments, for the
basic idea verification, we will use an SVM as the classifier. This
is to confirm that DML and the subsequent subspace mapping represent
the data in a form that benefits generic analysis. The basic DML algorithm
has been applied to a subset of about 3,500 samples of each of the
20 random datasets, where the sample size is reduced due to the high
computational cost of the straightforward optimisation of (\ref{eq:gobj}).
Of each subset, 80\% samples are used for training and validation
and 20\% for testing. Validation on the first subset shows that DML
is insensitive to the algorithm parameters, so the chosen parameter
set $k_{W}=10$, $k_{B}=20$ and $\beta=0.1$ is used throughout this
experiment (including following experiments of ADML on the full datasets).
Note that DML \emph{without} model selection in most (19 out of 20)
tests has a subtle \emph{beneficial} effect on the generality of the
results: our primary goal is to examine whether DML helps analysis
for a generic scenario; and thus the choice of SVM as the classifier
in the test stage should remain unknown to the DML algorithm in the
training stage as the data transformer, in order to prevent the DML
being ``overfit'' to SVM-oriented metrics. The parameters of SVM
have been determined by cross-validation in all tests. Fig. \ref{fig:dmlsvm}
(a) shows the classification performance by the first nearest neighbour
rule, SVM in the 3D data space and SVM in the DML-mapped 2D space.
From the figure, one can tell a meaningful improvement of classification
performance achieved by providing SVM the DML-mapped data over that
of applying SVM on the raw data. The results clearly show the DML-induced
subspace mapping does not only preserve useful information, but also
make the information more salient by suppressing irrelevant variance
through dimension reduction. Furthermore, in the DML-mapped space,
the superior classification performance has been achieved at a lower
computational cost. Fig. \ref{fig:dmlsvm} (b) displays the statistics
of the number of support vectors used by the SVMs in the raw 3D space
and the DML-mapped 2D space. In the reduced space, SVM can do with
a fraction of the support vectors that it has to use in the raw data
space for the same job.

In the second part of this experiment, we will examine our main idea
of the divide-aggregate implementation of DML, i.e. the ADML algorithm.
For the remaining tests in this subsection, each full dataset is split
into 50-50\% training and test subsets. The main focus is how the
division of the training data affects the learned distance metrics.

\begin{figure*}
\begin{centering}
\includegraphics[width=0.99\textwidth]{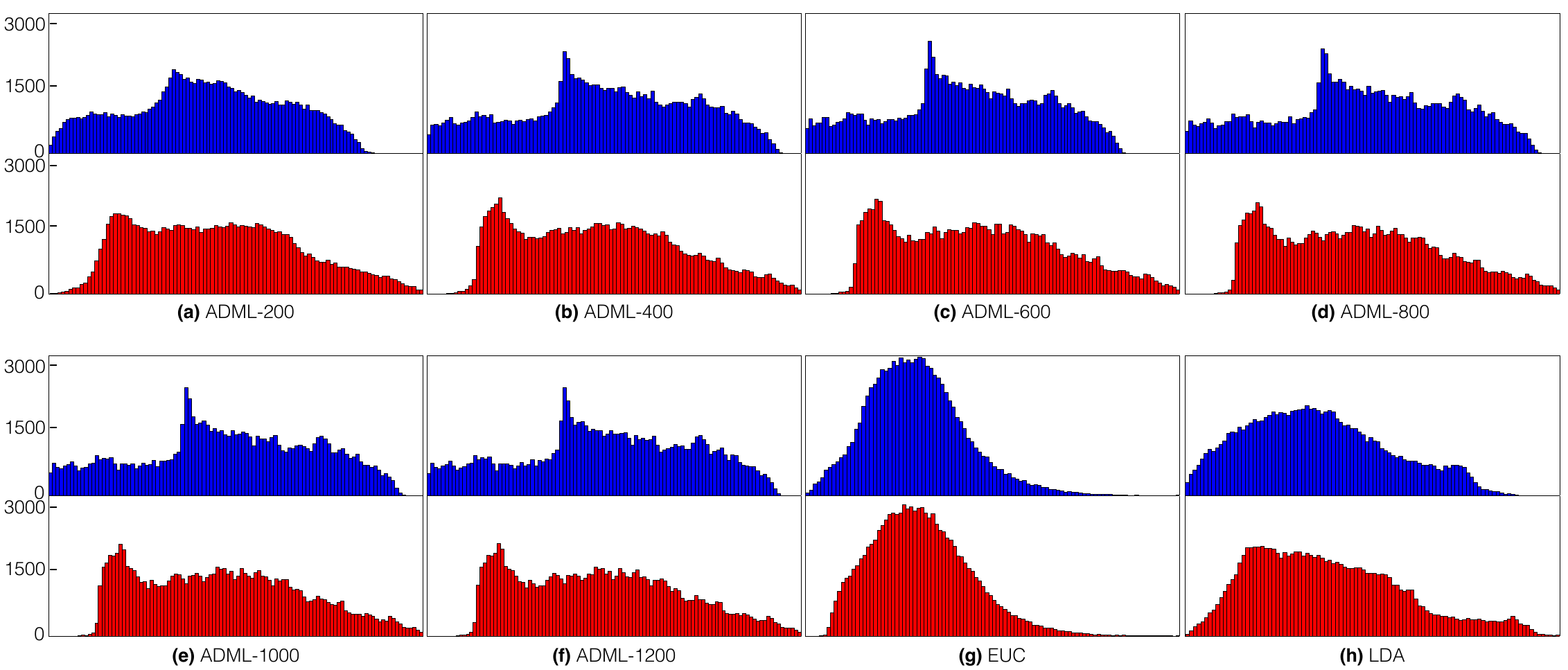}
\par\end{centering}

\protect\caption{Statistics of pair-wise distances of within- and between-class point
pairs. \label{fig:distcomp} Each subplot contains two histograms.
The upper (blue) histogram shows distances between 10,000 same-class
point pairs, normalised into {[}0,1{]}. Similarly, the lower (red)
histogram corresponds to cross-class pairs. Subplots \textbf{(a--f)}
display distance measured by ADML with different subdivisions of the
data (subset size ranging from 200 to 1,200). Subplot \textbf{(g)}
displays the histograms of Euclidean distance (EUC) and \textbf{(h)}
shows histograms of LDA-induced distance \textbf{(LDA)}.}
\end{figure*}

Fig. \ref{fig:distcomp} shows a close and explicit inspection of
how the learned distance metrics are connected to the class membership
of the data. We randomly sample 10,000 pairs of points, compute the
distance between the pair according to the learned metrics and normalise
the distances to $[0,1]$%
\footnote{Note each DML method has been run on 20 randomly generated datasets.
So 10,000 point pairs are drawn from each of those datasets and the
distance is computed w.r.t. the respectively learned metrics.%
}. Then two histograms are generated from the set of normalised distance,
one for those that both points in the pair are from the same class,
and another histogram corresponds to those that the pair of points
are from different classes. The two histograms are shown in the subplots
in Fig. \ref{fig:distcomp}, where the upper (blue) part represents
the same-class pairs and the lower (red) part represents the cross-class
pairs. A good metric should make the cross-class distance greater
than the same-class distance, especially for those smallest distances,
which correspond to nearest neighbours. The subplots in the figure
compare such statistical distinctions of several metrics. As shown
by the figure, DML leads to more informative distance metric than
the original Euclidean metric in the 3D space, which is consistent
with what we have observed in the earlier part of this experiment.
For comparison, the widely used linear discriminative analysis (LDA)
has also been used to project the data to 2D. Fig. \ref{fig:distcomp}
shows that LDA-induced distance metric make distinctions between the
same- and cross-class point pairs, but the difference is less than
that resulted by ADML-learned metric. A possible explanation is that
the same-class covariance of these datasets is close to the cross-class
covariance. Thus LDA, relying on the two covariance statistics, suffers
from the confusion between the two groups of covariance.

More importantly, the results shows ADML is reasonably stable over
the aggressive sub-divisions of the data. In particular, because ADML
trades accuracy for speed, ideally the sacrifice of performance would
be mild with regard to the partition of the data into smaller subsets
(lower costs). The results show that in these tests, the resultant
distance metric begins to help classification when the subset size
is as small as 200, and becomes distinctive between same- and cross-class
pairs when the subset size is greater than 400. Therefore the parallel
computation approximates the discrimination objective satisfactorily.

\begin{figure}
\begin{centering}
\includegraphics[width=0.48\textwidth]{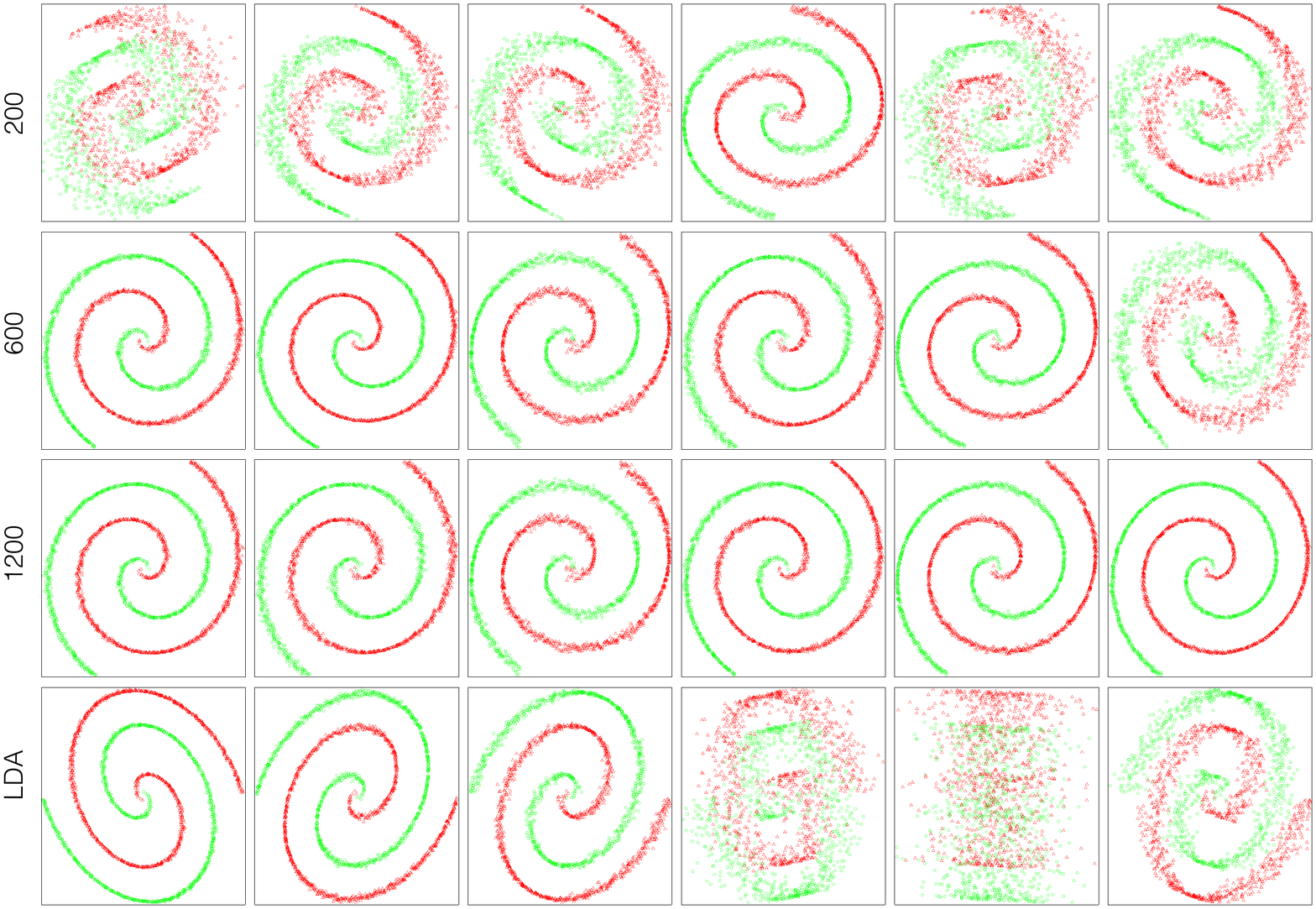}
\par\end{centering}

\protect\caption{\label{fig:toyproj} 2D projections induced by metric learning. The
figure shows 2D projections obtained by ADML (first three rows of
different subset sizes, 200, 600 and 1200) and LDA (last row) to project
six randomly generated datasets (columns). Projections distinguishing
samples from the two classes (green and red, better viewed in colours
on screen) are more preferred.}
\end{figure}

If a learned metric is represented by $\W$ and the data points are
$\X$, $\W^{T}\X$ represents a low dimensional projection. The data
points in the projection space reflect the geometry induced by the
learned metric. Fig. \ref{fig:toyproj} shows the projections of six
random datasets, where the projections are produced by ADML using
different data partitions, as well as by LDA. It is interesting to
see that given only 200 samples to individual local learners, ADML
starts producing projections that distinguish the two classes. With
larger sample sizes, the results are more reliable and the projections
become more discriminative. On the other hand, with all samples for
training, LDA performed unstably in these datasets, which may be due
to the large irrelevant component in the data covariance, as we have
discussed above.

\subsection{Automatic Image Annotation}

For a practical test, we apply metric learning algorithms to find
helpful distance measure for automatic annotation. Given a list of
interested concepts, the task of annotating an image is to determine
whether each of the concepts is present in the image. Essentially,
annotation is to establish a connection between the content of an
image and its appearance, which is known as the problem of semantic
gap \cite{Sme00}. A natural solution to the problem is to identify
the image in question with respect to a set of reference images that
have been labelled with interested tags of concepts. Therefore, given
the features of visual appearance, successful annotation relies on
finding useful geometry in the appearance space, which reflects the
connection between images in the semantic space; and annotation will
benefit from effective metric learning methods. In the following report,
we discuss the details of applying metrics learned by different techniques
to annotate a large scale image dataset.

\subsubsection{Task specification and evaluation criterion \label{sub:imganno-spec}}

The NUS-WIDE dataset %
\footnote{\url{http://lms.comp.nus.edu.sg/research/NUS-WIDE.htm}%
} \cite{Chu09} has been used for the annotation test. The dataset
has been compiled by gathering publicly available images from Flickr.
To facilitate comparison, the visual appearance of each image is represented
by six sets of standard features. These features are: colour histogram
(of 64 bins), colour correlogram (of 144 bins), edge direction histogram
(of 73 bins), wavelet texture (of 128 coefficients), block-wise colour
moments (of 255 elements), and bag-of-words feature based on the SIFT
descriptors (of 500 bags). There are totally 1134 visual features
for each image. The raw data are centred at zero, and each feature
is normalised to have unit variance. The task is to choose from 81
tags for each image. The tags are determined independently, thus each
image can have more than one tag. 

The annotation procedure is based on the nearest neighbour rule. To
decide the contents of an image, up to 15 of its nearest neighbours
are retrieved according to the tested distance metric. We compare
the level of presence of the tags among the nearest neighbours of
an image with their background levels in the reference dataset, and
predict the tags of the test image accordingly. If a tag is assigned
to $r_{0}$ percent of all images in the reference dataset, and is
assigned to $r_{1}$ percent of images among the nearest neighbours,
the tag is predicted to be present in the test image if $r_{1}>r_{0}$.
The number of nearest neighbours used for predicting the tags has
been determined by cross validation in our tests (from 1 to 15). The
nearest neighbour rule has been shown effective for the annotation
task \cite{Jeo03}. More importantly, compared to other (multi-label)
classification methods, the nearest neighbour rule provides the most
direct assessment of the learned distance metrics using different
techniques. The annotation reflects how the semantic information in
the reference set is preserved in the learned distance metric in the
appearance feature space. 

F1-scores is measured for the prediction performance of each tag among
the test images. The average value of all the tags is reported as
a quantitative criterion. Having predicted the tags for a set of test
images, the F1-score is measured as suggested in \cite{Rij79}, with
the following combination of \emph{precision} and \emph{recall} 
\begin{align}
F1 & =\frac{2\times\textrm{precision}\times\textrm{recall}}{\textrm{precision}+\textrm{recall}}\label{eq:f1}
\end{align}
where precision and recall are two ratios, ``correct positive prediction
to all positive prediction'' and ``correct positive prediction to
all positive''. F1 score ranges from 0 to 1, higher scores mean better
result. 

In the experiment, ADML and state-of-the-art distance metric learning
methods are tested, as well as several baseline metrics particularly
designed for image annotation. These distance metrics are: 
\begin{itemize}
\item Baseline, Euclidean (Base-EUC): the basic baseline metric is the Euclidean
distance in the space of raw features (normalised to zero-mean and
unit-variance). This metric does not require training.
\item Baseline, Joint Equal Contribution (Base-JEC): suggested by \cite{Makadia08},
this baseline attributes equal weights to the six different types
of image features. This metric does not require training.
\item Baseline, $L_{1}$-penalised Logistic Regression (Base-LR): also suggested
by \cite{Makadia08}, this baseline attributes $L_{1}$-penalised
weights to the six types of image features to maximise dissimilarity
and minimise similarity according to given tags. This metric is requires
adaptation of the weights. However, after preprocessing, the penalised
regression problem is basic and without altering the Euclidean metric
of individual feature set. Thus we do not consider the training cost.
\item Tag Propagation (TagProp): suggested by \cite{tagprop}, a specialised
image annotation method. This technique weighs the six feature sets
by considering the nearest neighbours of individual images. 
\item Distance metric learning methods discussed in Section \ref{sec:rev}
including Xing et al.'s method (Xing) \cite{Xin03}, Relevance Component
Analysis (RCA) \cite{She02}, Discriminative Component Analysis (DCA)
\cite{Hoi06}, Large Margin Nearest Neighbour Classifier (LMNN) \cite{Wei06}
and Information-Theoretic Metric Learning (ITML) \cite{Dav07}.
\end{itemize}
For Xing, RCA, DCA, ITML and ADML, the supervision is in the form
of similar/dissimilar constraints. In the tests, 100,000 image pairs
are randomly sampled from the dataset to measure the ``background''
number of shared concepts between images. Then a pair of image is
considered similar if they share more concepts (common tags) than
the background number, or dissimilar otherwise.

In all the tests, algorithm parameters are chosen by cross validation.
For ADML, the important settings are as follows. The number of columns
of matrix $\W$ varies from $50$ to $100$ (subspace dimension).
The within-class neighbourhood size $k_{1}$ varies from $1$ to $20$
and the between-class neighbourhood size $k_{2}$ varies from $20$
to $80$. The coefficient $\beta$ has been chosen from $0.1$ to
$2.0$. We have tried different sizes for the subsets from $200$
to $1,000$.

\subsubsection{Test on medium-sized subsets \label{sub:subset}}

We first access the learned and baseline metrics with respect to different
sizes of data and different levels of complexity of the annotation
task. Four medium-sized subsets of the NUSWIDE datasets are compiled.
Each subset consists of images with a small number (2-5) of labels.
In particular, we take the two most popular concepts in the dataset,
then let Subset I contain images with tags of at least one of the
two concepts. Subset II subsumes Subset I, and includes extra images
of the third most popular concepts. The construction carries on until
we have Subset IV consisting of images with the five most popular
concepts. Of all the samples in each subset, 60\% of the samples are
used for training and 40\% for test. Among the training samples, about
15\% are withheld for cross validation to fix algorithm parameters.

\begin{table*}
\protect\caption{Annotation Performances and Time Costs on Subsets I-IV of the NUS-WIDE
Dataset.\label{tab:subset} }

\begin{centering}
\includegraphics[width=16cm]{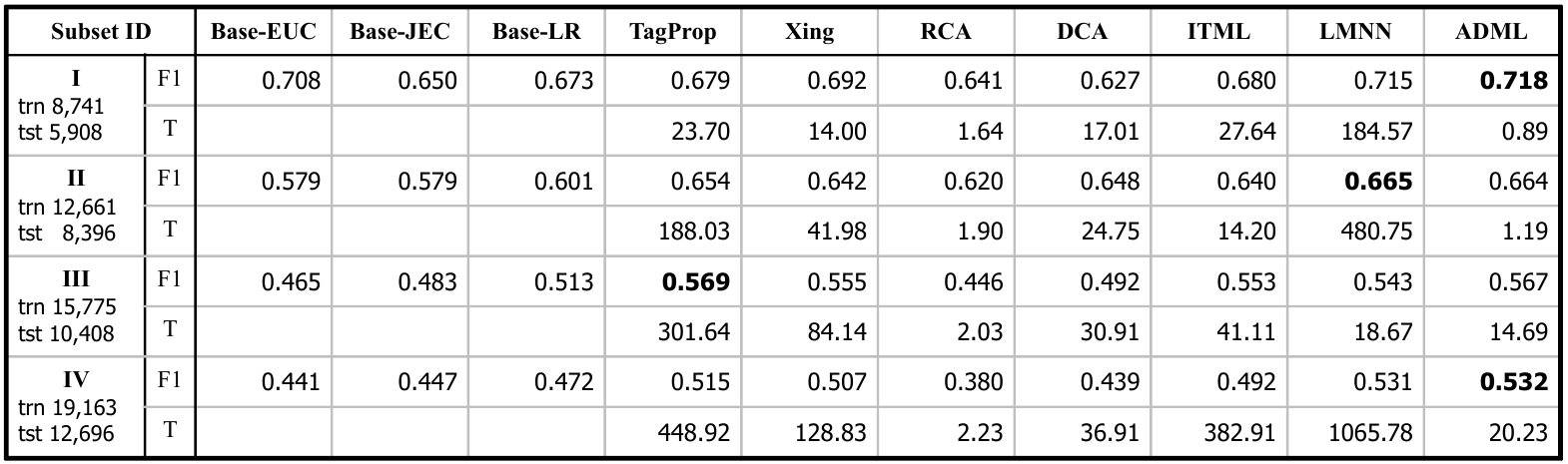}
\par\end{centering}

Each column represents a measure of distance. Each subset corresponds
to a pair of rows, where ``trn/tst'' represents the number of training
(including validation) and test samples, \textquotedbl{}F1\textquotedbl{}:
the average F1-scores of predicting the labels as defined in (\ref{eq:f1}),
\textquotedbl{}T\textquotedbl{}: the \emph{wall} time of the \emph{training}
stage. Detailed explanation of the results and further analysis are
provided in Subsection \ref{sub:subset}.
\end{table*}

The annotation and evaluation processes follow the discussion above.
The average F1-scores achieved by each metric in each subset are listed
in Table \ref{tab:subset}. The results show that accuracy of predicting
the concepts is affected by the distance metrics. When the task becomes
more complex (to predict more concepts), the learned metrics are increasingly
more helpful than the baseline metrics derived from the standard Euclidean
distance. More specifically, in these tests, the more effective learning
techniques are those considering the local discriminative geometry
of individual samples, such as ITML, LMNN and ADML, compared to those
working on the global data distribution, such as DCA and RCA. In all
the tests, ADML has provided a distance metric achieving superior
or close-to-the-best annotation performance. 

More relevant to our main motivation of the efficiency of metric learning,
Table \ref{tab:subset} lists the time cost of the tested techniques.
We report the \emph{wall} \emph{time} elapsed during the training
stage of the algorithms. In addition to theoretical computational
complexity, wall time also provides a practical guide of the time
cost. This is important because the parallel processing of ADML needs
a comprehensive assessment including the communication time and other
technical overheads, which is difficult to gauge using only the CPU
time of the core computations. The discussion in \cite{Blelloch96}
provides a more comprehensive view of complexity in the light of parallel
computation.

The reported time has been recorded on a computer with the following
configuration. The hardware settings are 2.9GHz Intel Xeon E5-2690
(8 Cores), with 20MB L3 Cache 8GT/s QPI (Max Turbo Freq. 3.8GHz, Min
3.3GHz) and 32GB 1600MHz ECC DDR3-RAM (Quad Channel). We have used
the Matlab implementation of the metric learning algorithms (Xing,
RCA, DCA, ITML and LMNN) provided by their authors. To facilitate
comparison, we implement ADML in Matlab as well, with the help of
Parallel Computation Toolbox. The algorithms are run on Matlab 2012b
with six workers. 

In the tests, ADML has achieved good performance with high efficiency.
Moreover, at the algorithmic level, the relative advantage of ADML
could be greater than that is shown by the practical results in Table
\ref{tab:subset}. First, all algorithms are tested on the same Matlab
environment, which in the background invokes Intel Math Kernel Library
(MKL) for the fundamental mathematical libraries. MKL accelerates
all algorithms, however, MKL competes with ADML's parallel local metric
learners for limited cores on a machine. It can be expected that ADML
will benefit greater from increasing number of processors than the
rival algorithms. This will also apply if advanced computational hardware
is utilised such as GPU- or FPGA-based implementations\cite{fpga,Rai09}.
Second, multiple subset sizes (from 200 to 1,000) have been explored
for ADML, and the reported performance is the one that performed best
on the validation set. However, as we have shown in the tests on the
synthetic dataset in Subsection \ref{sub:toy}, the size of the subsets
has limited effect on the final metric. We will discuss this issue
in more details later in the next experiment on the full dataset.

\subsubsection{Test on the full dataset \label{sub:full}}

\begin{table}
\protect\caption{Annotation of NUS-WIDE Dataset\label{tbl:fullNUS}}

\centering{}\includegraphics[width=6.6cm]{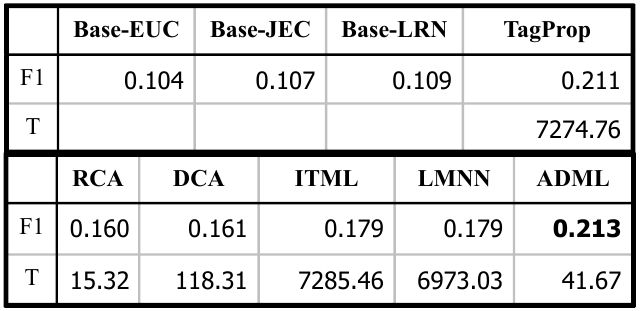}
\end{table}

The annotation test has been conducted on all the images in the NUS-WIDE
dataset that are labelled with at least one concept. As above, 60\%
images are used for training and 40\% are used for test, giving $79,809$
training images and $53,632$ test images. Of the training images,
15\% are used for cross validation to choose algorithm parameters.
There are 81 concepts to be labelled in this dataset.

Table \ref{tbl:fullNUS} shows the annotation performances of the
tested metrics. There are similar trends as those obtained on the
four subsets in the last experiment. With the complexity of predicting
81 concepts, the advantage of learned metrics over the baseline metrics
is more significant. ADML yields superior metrics using less time
compared to the rival methods.

\begin{figure}
\begin{centering}
\includegraphics[width=3in]{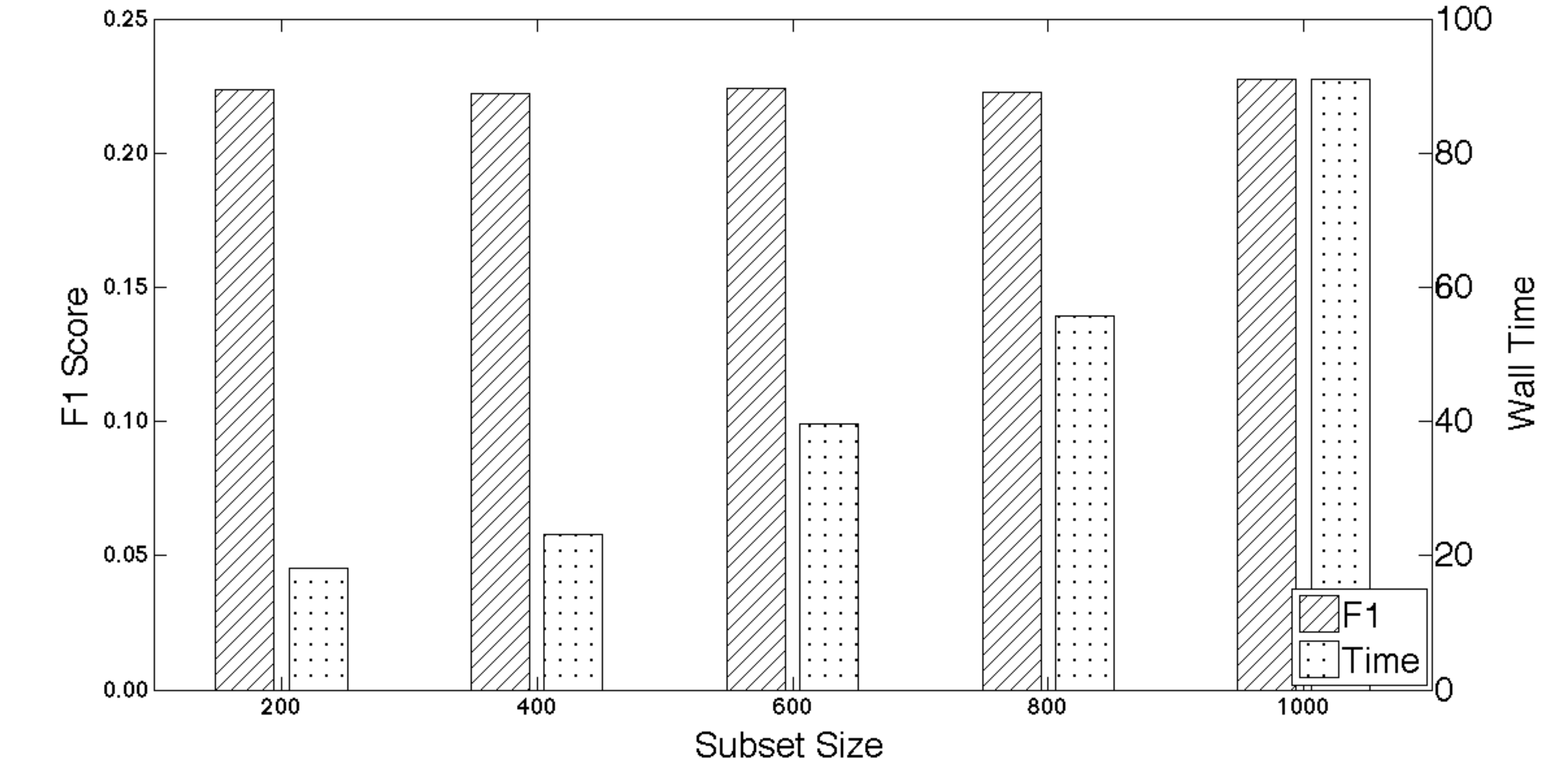}
\par\end{centering}

\protect\caption{\label{fig:ss}Annotation Performance and Time Cost w.r.t. Subset
Sizes}
\end{figure}

In the experiments above, we choose the parameters for ADML by cross
validation and configure the algorithm for optimal annotation performance.
On the other hand, the primary motivation for ADML is to save the
time cost of metric learning, and the setting of the subset size is
a major factor of the overall time cost. Thus it is useful to study
how ADML behaves under varying sizes of the subsets. Fig. \ref{fig:ss}
shows the F1-scores using the metrics learned with different subset
sizes. The figure also compares the wall time cost in the training
processes. The experiment demonstrates that ADML is stable w.r.t.
different data partitions. Thus the algorithm can achieve desirable
time efficiency with little sacrifice in the quality of learned distance
metrics. Note that this conclusion corroborates the result on the
synthetic data shown in Fig. \ref{fig:distcomp}.

\begin{figure}
\centering{}\includegraphics[width=0.45\textwidth]{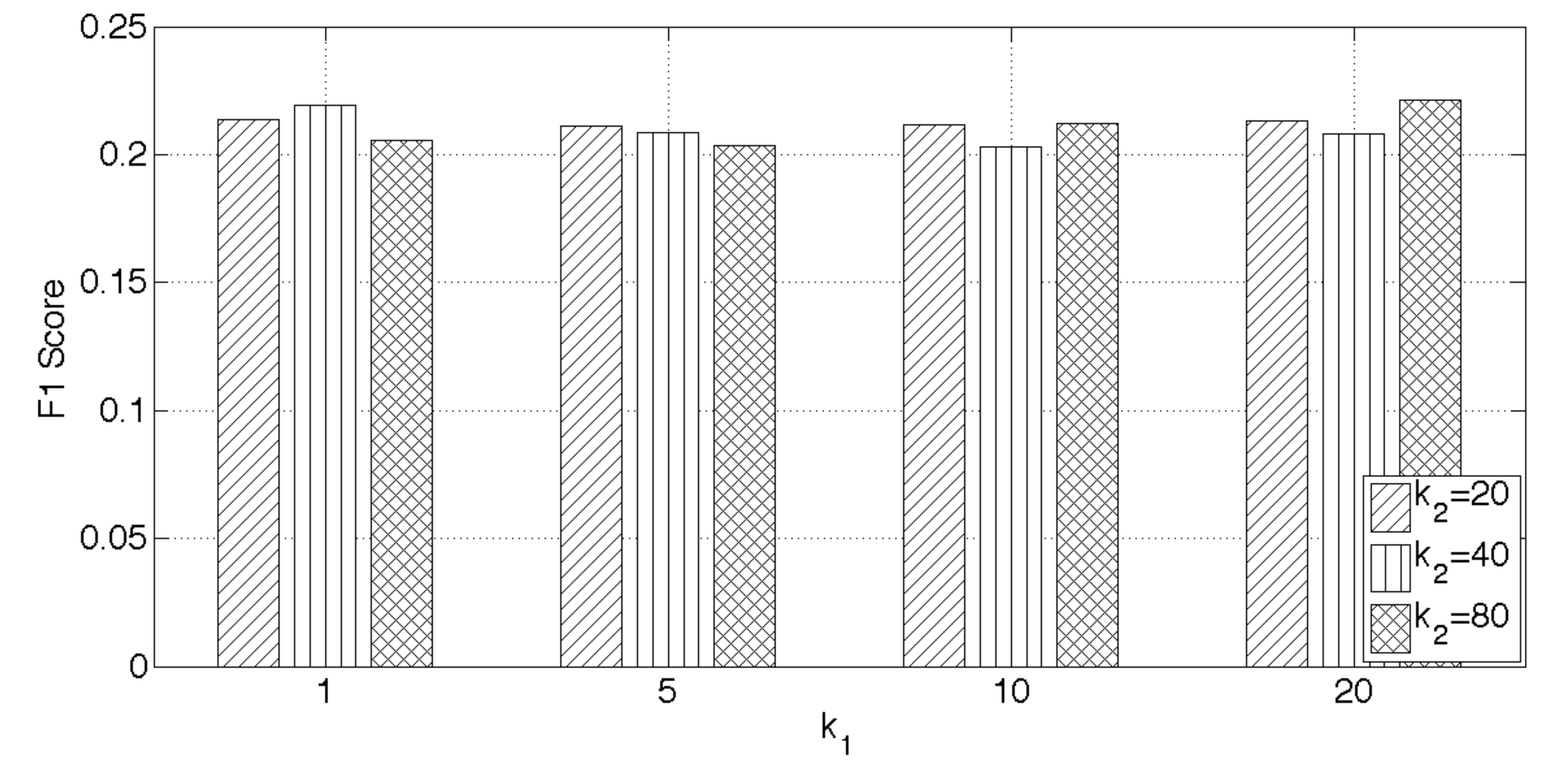}\protect\caption{\label{fig:k} Annotation Performance w.r.t. Within- and Between-class
Neighbourhoods}
\end{figure}

\begin{figure}
\begin{centering}
\includegraphics[width=0.45\textwidth]{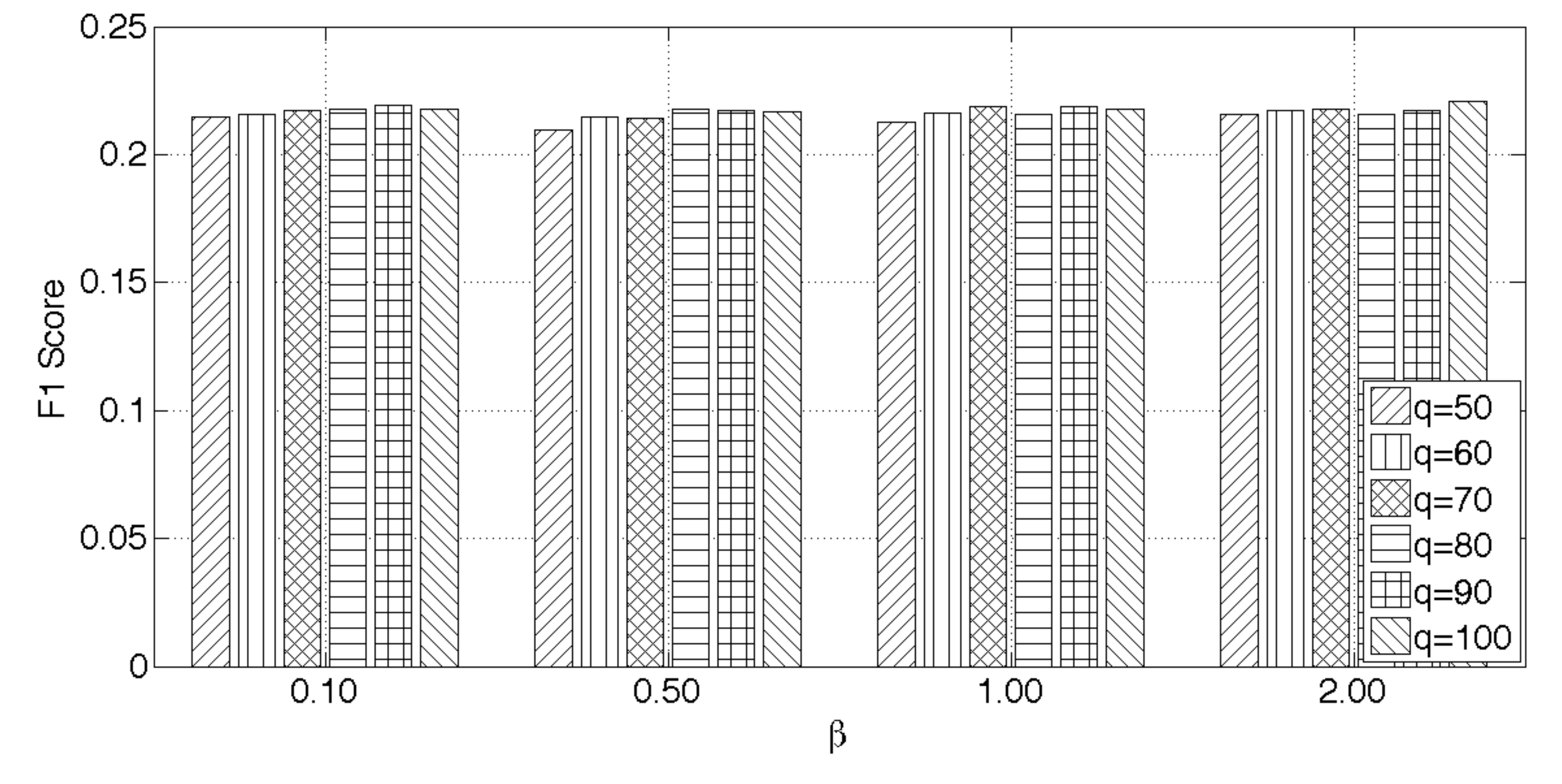}
\par\end{centering}

\protect\caption{\label{fig:qbt} Annotation Performance w.r.t. $\beta$ and subspace
dimension}
\end{figure}

In our experiment, the performance of ADML is relatively stable with
respect to the other algorithm settings, including the size of within-
and between-class neighbourhoods $k_{1}$ and $k_{2}$, the weight
$\beta$ and the subspace dimension $q$ (refer to Section \ref{sec:DDML}
for detailed explanations of the parameters). Fig. \ref{fig:k} and
Fig. \ref{fig:qbt} show the annotation performance of ADML with varying
$k_{1}$, $k_{2}$, $\beta$ and $q$. Note the adjustment of the
parameters are grouped into two pairs, $(k_{1},k_{2})$ and $(\beta,q)$,
so that we report more behaviours of ADML using less space.

\section{CONCLUSION\label{sec:conclu}}

In this paper, we have proposed a distance metric learning approach
that emphasises on encoding categorical information of data. We further
derive a part-based solution of the algorithm: the metrics are learned
on subsets divided from the entire dataset, and those partial solutions
are aggregated into a final distance metric. The aggregated distance
metric learning (ADML) technique takes advantage of distributed and
parallel computation, and makes metric learning efficient in terms
of both time and space usage. 

To justify the gain in efficiency, we provide support for the learning
quality of ADML in both the theoretical and practical aspects. Theoretically,
error bounds are proved showing that the divide-and-conquer technique
will give results close to that would be resulted by the performing
the discriminative learning method on the entire dataset. Empirically,
the properties of the metric learned by ADML have been shown helpful
in reflecting the intrinsic relations in the data and helping practical
image annotation task. The success of the partition-based processing
may also be explained by the theory about the bias-variance trade-off
\cite{Has03}. Learning on a subset, as opposed to the entire data
set, may introduce extra bias; on the other hand, ADML can be seen
as the weighted combination of results from each subset, which will
cause the decrease of variance. Thus, the overall performance may
not be affected seriously. Theoretical exploration in this direction
make an interesting research subject.

It worth noting that in our empirical study, ADML is implemented using
Matlab and the Parallel Computation Toolbox. This implementation facilitates
the comparison with other metric learning techniques, but as discussed
above, the parallel computation in ADML overlaps those inherited from
the Intel's MKL fundamental subroutines (utilised by Matlab's low-level
operations). ADML can be implemented full-fledged MapReduce architecture,
which enables the algorithm to scale up to very large scale and deal
with data across multiple nodes of high-performance clusters. A sketch
of a distributed implementation is given in Algorithm \ref{alg:adml}.
The algorithm has been tested practically on a cluster. Via OpenMPI,
we distributed 32 processes across 8 nodes of Intel Xeon E5 series,
each having 8 cores of 2.9 or 3.4GHz%
\footnote{In particular, to avoid irrelevant variables in a complex multi-user
environment, we selected machines of light load at the time of testing
(<1\%), and utilised only 50\% of the cores available. The master
process communicated with 31 workers via an intranet.%
}. The data are sampled from the distribution as discussed in Subsection
\ref{sub:toy}. The master process in Algorithm \ref{alg:adml} learned
metrics from more than 188M samples in 855s, where it aggregated 200k
local learned metrics from 31 worker process. In contrast, a sequential
implementation of the basic algorithm in Section \ref{sec:DDML} took
1718s to learn from 1.88M samples, 1\% of the data tackled by the
parallel Algorithm \ref{alg:adml}.

{\small{}}
\begin{algorithm}
\begin{lyxcode}
\noindent {\small{}\includegraphics[width=6cm]{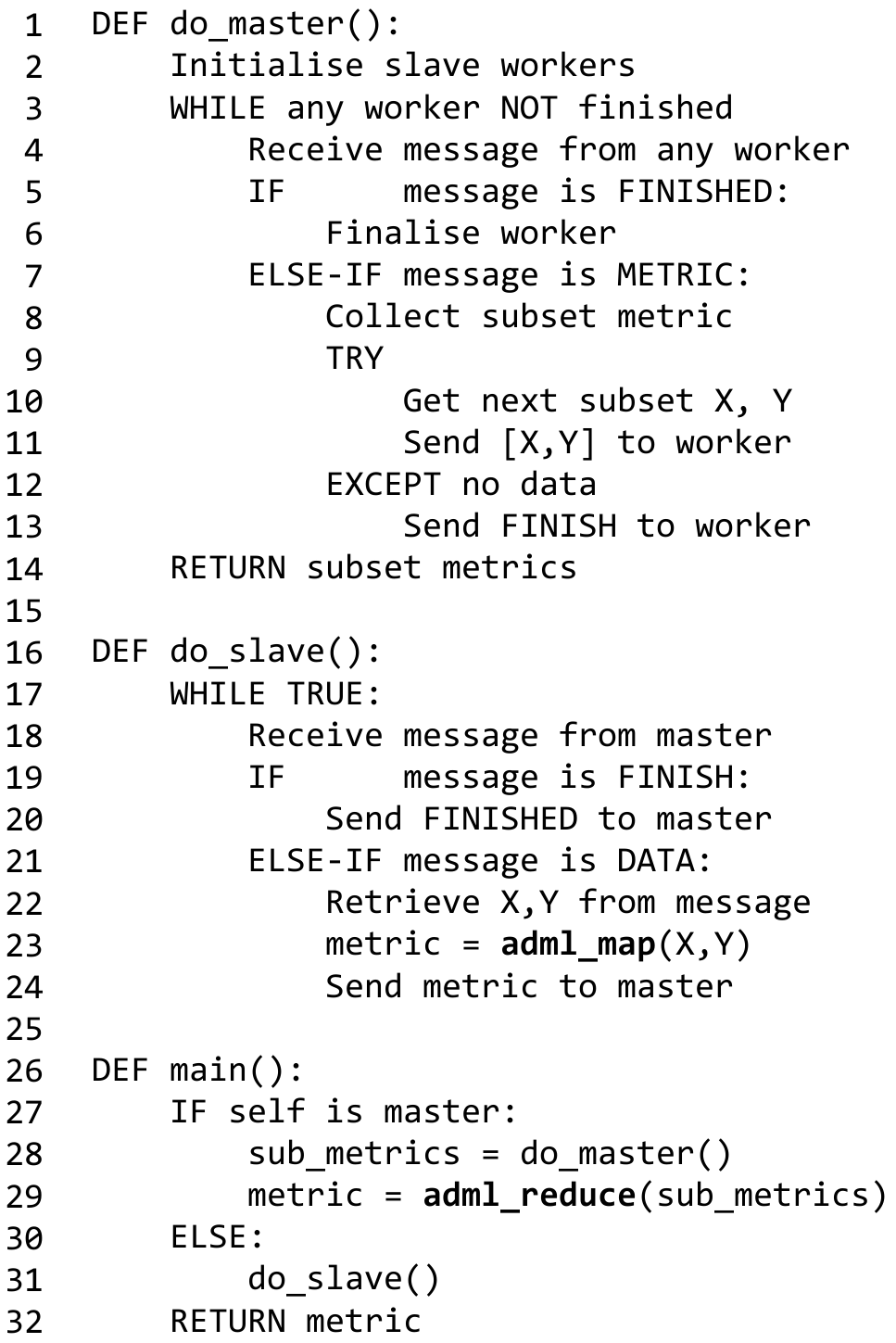}}{\small \par}
\end{lyxcode}
{\small{}The map-steps, ``adml\_map()'', compute subset metrics
according to (\ref{eq:W.obj}) (See Section \ref{sec:ADML}). The
``adml\_reduce'' does the aggregation according to (\ref{eq:agg-by-svd}).
The pseudo code is assuming a Message Passing Interface (MPI) as the
MapReduce architecture. It is easily convertible to other major programming
languages and MapReduce realisation such as Java-Hadoop. Note that
for a completely distributed implementation, the procedure can be
further decentralised by adjusting the data-accessing in Line 10.
The master process may send instructions of retrieving locally stored
data, instead of loading the data and sending them to the worker processes.}{\small \par}

{\small{}\protect\caption{Pseudo code of implementing ADML in a MapReduce framework\label{alg:adml}.}
}
\end{algorithm}
{\small \par}

\appendix{}

\subsubsection*{Theorem \ref{the:3.1}}
\begin{IEEEproof}
Part I: If $\R=\sum_{k}\R_{k}$ is invertible, we are allowed to write
$\hat{\W}=(\sum_{k}\R_{k})^{-1}(\sum_{k}\R_{k})\hat{\W}$. Subtracting
$\hat{\W}$ from the aggregation rule (\ref{eq:SolTotalProjMat})
gives
\begin{align*}
\W_{A}-\hat{\W}= & {\textstyle \R^{-1}\sum_{k=1}^{K}\SubSamLapk\left(\W_{k}-\hat{\W}\right).}
\end{align*}
Matrix triangle inequality \cite{Hor90} states that $\|\mat A_{1}+\mat A_{2}+\dots\|\leq\|\mat A_{1}\|+\|\mat A_{2}\|+\dots$,
thus 
\begin{align}
\bigl\Vert\W_{A}-\hat{\W}\bigr\Vert & ={\textstyle \Bigl\Vert\R^{-1}\Bigl(\sum_{k=1}^{K}\SubSamLapk\bigl(\SubProjMatk-\hat{\W}\bigr)\Bigr)\Bigr\Vert}\nonumber \\
 & \!\!\!\!\!\!\!\!\!\!\!\!\!\!\!\!\!\!\!\!\!\!\!\!\!\!\leq{\textstyle \sum_{k=1}\bigl\Vert\R^{-1}\SubSamLapk\bigl(\SubProjMatk-\hat{\W}\bigr)\bigr\Vert}\nonumber \\
 & \!\!\!\!\!\!\!\!\!\!\!\!\!\!\!\!\!\!\!\!\!\!\!\!\!\!\le{\textstyle \max_{k}\bigl\Vert\SubProjMatk-\hat{\W}\bigr\Vert\times\bigl\Vert\R^{-1}\bigr\Vert{\textstyle \sum_{k=1}^{K}}\left\Vert \SubSamLapk\right\Vert }.\label{eq:pf1-part1-a}
\end{align}
Since for a symmetric matrix $\mat A$, $\|\mat A\|=\sqrt{\lambda_{max}(\mat A^{T}\mat A)}=\lambda_{max}(\mat A)\geq\lambda_{min}(\mat A)$,
\begin{align}
{\textstyle \left\Vert \R^{-1}\right\Vert \sum_{k=1}^{K}\left\Vert \SubSamLapk\right\Vert } & \le{\textstyle \frac{1}{\lambda_{min}(\R)}\sum_{k}\lambda_{max}(\R_{k})}\nonumber \\
 & {\textstyle \le\frac{K\max_{k}\{\lambda_{max}(\R_{k})\}}{\lambda_{min}(\R)}}=S_{1}.\label{eq:pf1-part1-b}
\end{align}
Combining (\ref{eq:pf1-part1-a}) and (\ref{eq:pf1-part1-b}), we
have the first part of Theorem \ref{the:3.1}. 
\end{IEEEproof}
For the second part, we will need the following lemma.
\begin{lem}
\label{lem:sum-norm} If $\A=\A_{1}+\A_{2}+\dots+\A_{K}$, and $\{\A_{k}\}$
are positive definite matrices, then $\|\A\|\ge K\min_{k}\{\lambda_{min}(\A_{k})\}$.\end{lem}
\begin{IEEEproof}
As the sum of positive definite matrices, $\A$ is a positive definite
matrix itself. For a positive definite matrix, 
\begin{align*}
\A^{T}\A\v & =\lambda\v\Leftrightarrow\A\v=\sqrt{\lambda}\v\\
\textrm{and }\|\A\| & =\lambda_{max}(\A)=\max_{\v}\left\{ \frac{\v^{T}\A\v}{\|\v\|^{2}}\right\} .
\end{align*}
Thus it is sufficient to show there exists some $\v$ such that
\begin{align*}
\v^{T}\A\v & \ge K\min_{k}\{\lambda_{min}(\A_{k})\}\times\|\v\|^{2}.
\end{align*}
Since for any $\v$, $\v^{T}\A_{k}\v\ge\lambda_{min}(\A_{k})$, we
have
\begin{align*}
\v^{T}\A\v & =\sum_{k}\v^{T}\A_{k}\v\ge\sum_{k}\lambda_{min}(\A_{k})\|\v\|^{2}\\
 & \ge K\min_{k}\{\lambda_{min}(\A_{k})\}\times\|\v\|^{2}
\end{align*}

\end{IEEEproof}
Lemma \ref{lem:sum-norm} leads to the second part of Theorem \ref{the:3.1}:
since $\|\R\|\ge K\min_{k}\{\lambda_{min}(\R_{k})\}$, we can combine
$\|\R\|^{-1}\le\frac{1}{K\min_{k}\{\lambda_{min}(\R_{k})\}}$ with
(\ref{eq:pf1-part1-b}) and arrive the desired conclusion.

\subsubsection*{Theorem \ref{the:alter-agg-consis}}
\begin{IEEEproof}
The aggregation rule (\ref{eq:agg-by-svd}) indicates
\begin{align}
\W_{A} & =\sum_{k=1}^{K}\SubSamLapk\SubProjMatk\V\S^{-1}.\label{eq:agg-by-svd-explicit}
\end{align}
Subtracting $\hat{\W}$ from both sides of (\ref{eq:agg-by-svd-explicit})
gives{\small{}
\begin{align}
 & \|\W_{A}-\hat{\W}\|\nonumber \\
= & {\textstyle \|\sum_{k=1}^{K}\SubSamLapk\SubProjMatk\V\S^{-1}-\hat{\W}\|}\nonumber \\
\leq & {\textstyle \|\sum_{k}\R_{k}(\W_{k}-\hat{\W})\V\S^{-1}\|+\|\sum_{k}\R_{k}\hat{\W}\V\S^{-1}-\hat{\W}\|}\label{eq:mi1}\\
\leq & {\textstyle \|\R\|\bigl\Vert\sum_{k}(\W_{k}-\hat{\W})\V\S^{-1}\bigr\Vert+\|\R\hat{\W}\V\S^{-1}-\hat{\W}\|}\label{eq:mi2}\\
\le & {\textstyle \frac{\lambda_{max}(\R)}{\min\{\textrm{diag}(\S)\}}\left(1+\sum_{k}\|\W_{k}-\hat{\W}\|\right)+1,}\label{eq:th2prv}
\end{align}
}where (\ref{eq:mi1}) and (\ref{eq:mi2}) are by the the matrix triangle
inequality. The relation (\ref{eq:th2prv}) translates directly to
Theorem \ref{the:alter-agg-consis}.\end{IEEEproof}

\bibliographystyle{IEEEtran}
\bibliography{biblo}

\begin{IEEEbiography}[{\includegraphics[width=1in,height=1.25in,clip,keepaspectratio]{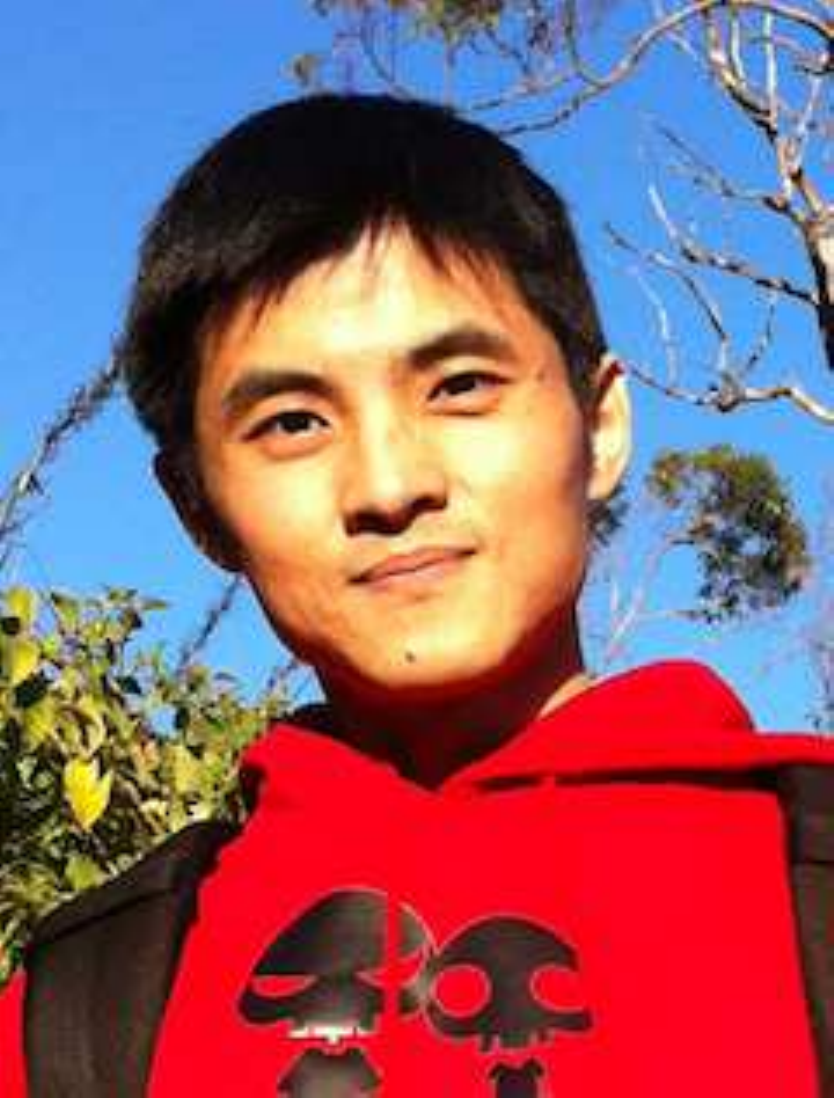}}]{Jun Li}
received his BS degree in computer science and technology from Shandong
University, Jinan, China, in 2003, the MSc degree in information and signal
processing from Peking University, Beijing, China, in 2006, and the PhD degree
in computer science from Queen Mary, University of London, London, UK in 2009.
He is currently a research fellow with with the Centre for Quantum Computation
and Information Systems and the Faculty of Engineering and Information
Technology in the University of Technology, Sydney.
\end{IEEEbiography}

\begin{IEEEbiography}[{\includegraphics[width=1in,height=1.25in,clip,keepaspectratio]{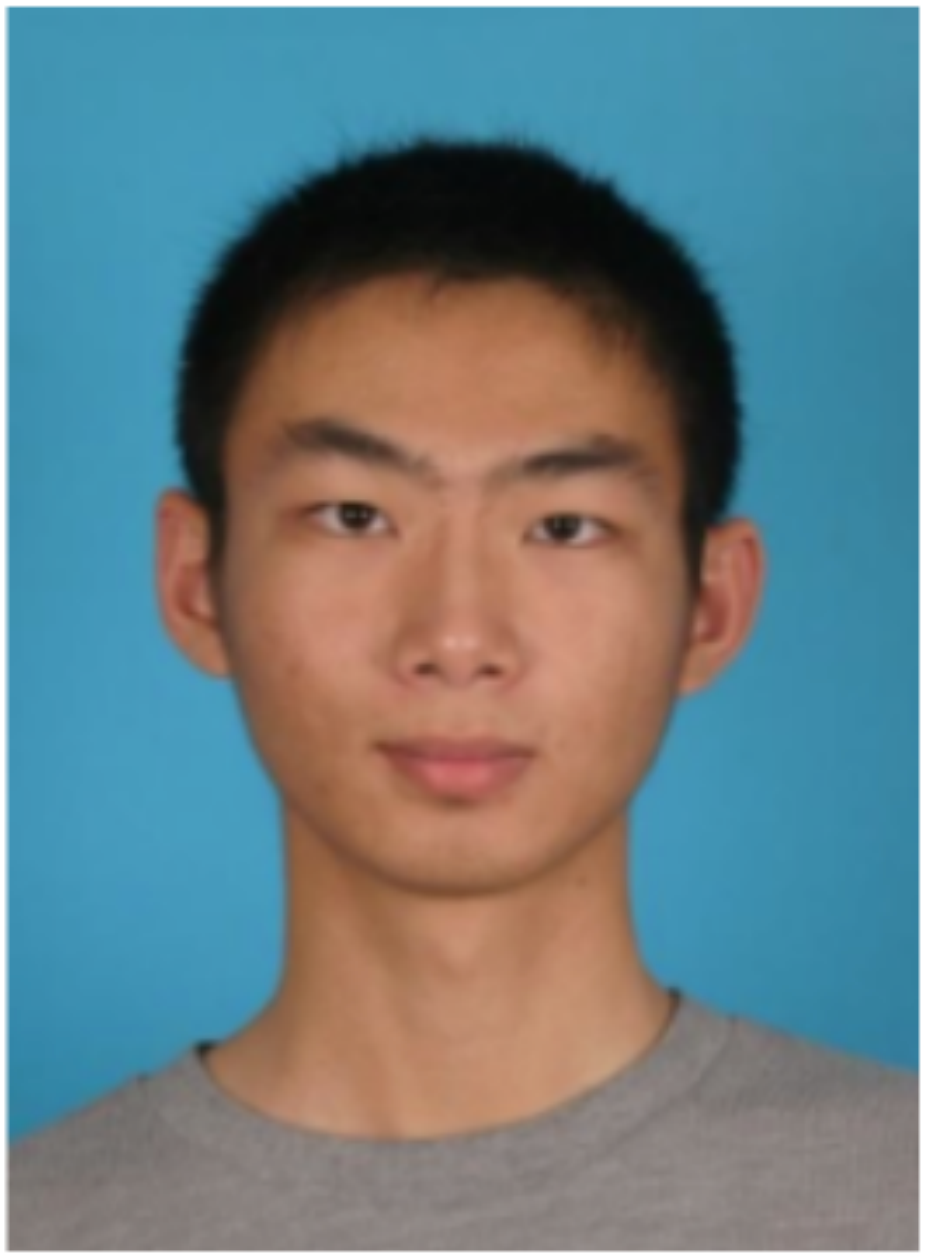}}]{Xu Lin}
received a BEng degree from South China University of Technology. He is
currently with the Shenzhen Institutes of Advanced Technology, Chinese Academy
of Sciences, Shenzhen, China, as a visiting student for Human Machine Control.
Over the past years, his research interests include machine learning, computer
vision.
\end{IEEEbiography}

\begin{IEEEbiography}[{\includegraphics[width=1in,height=1.25in,clip,keepaspectratio]{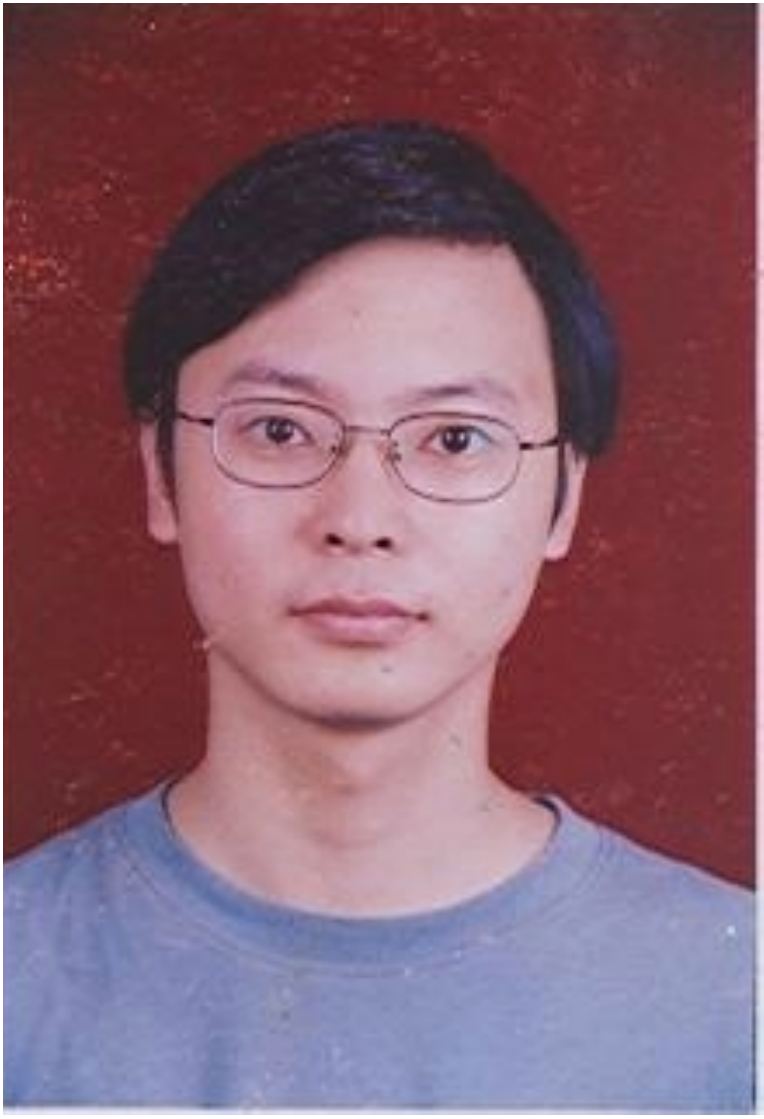}}]{Xiaoguang Rui}
has played the key roles in many high impact projects in IBM CRL
for industry solutions, and published several papers and filed
disclosures/patents.  He has received the master and doctoral degrees
consecutively in the Department of Electronic Engineering and Information
Science (EEIS), University of Science and Technology of China (USTC). His
research interests include data mining, and machine
learning.
\end{IEEEbiography}

\begin{IEEEbiography}[{\includegraphics[width=1in,height=1.25in,clip,keepaspectratio]{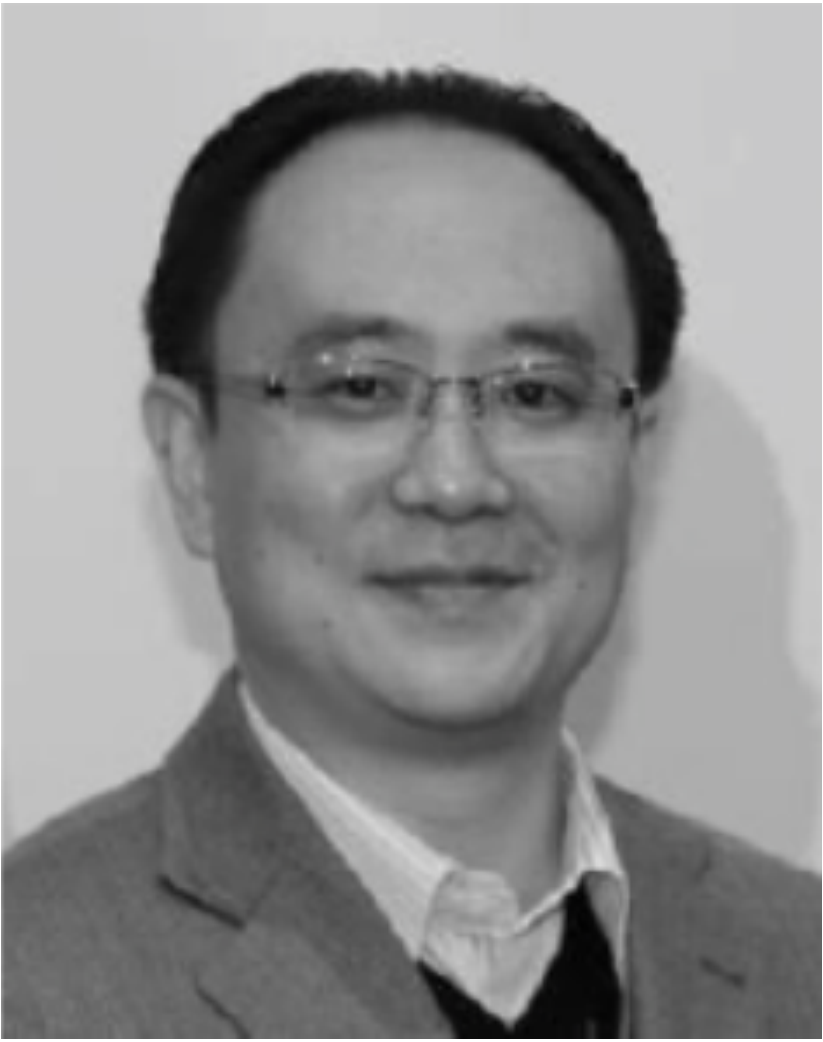}}]{Yong Rui}
(F'10) is currently a Senior Director and Principal Researcher at Microsoft
Research Asia, leading research effort in the areas of multimedia search and
mining, knowledge mining, and social computing. From 2010 to 2012, he was
Senior Director at Microsoft Asia-Pacific R\&D (ARD) Group in charge of China
Innovation. From 2008 to 2010, he was Director of Microsoft Education Product
in China. From 2006 to 2008, he was a founding member of ARD and its first
Director of Strategy. From 1999 to 2006, he was Researcher and Senior
Researcher of the Multimedia Collaboration group at Microsoft Research,
Redmond, USA.
A Fellow of IEEE, IAPR and SPIE, and a Distinguished Scientist of ACM, Rui is
recognized as a leading expert in his research areas. He holds 56 issued US and
international patents. He has published 16 books and book chapters, and 100+
referred journal and conference papers.
\end{IEEEbiography}

\begin{IEEEbiography}[{\includegraphics[width=1in,height=1.25in,clip,keepaspectratio]{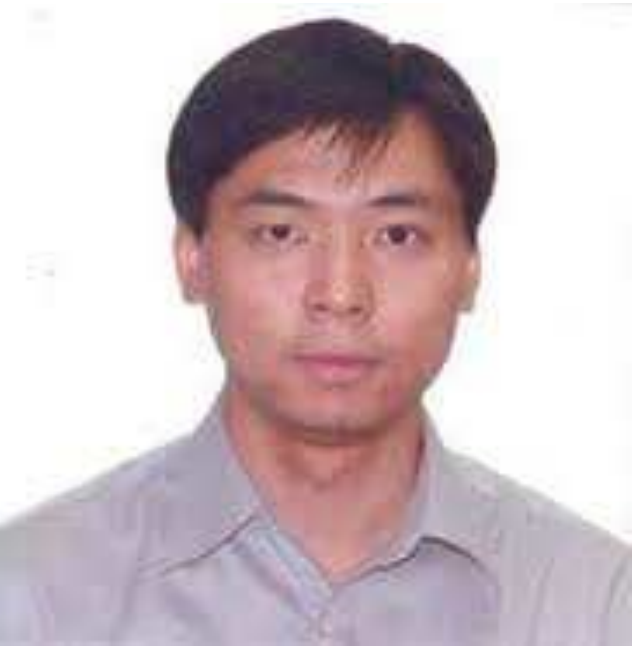}}]{Dacheng Tao}
(M'07-SM'12) is Professor of Computer Science with the Centre for
Quantum Computation and Intelligent Systems and the Faculty of Engineering and
Information Technology in the University of Technology, Sydney. He mainly
applies statistics and mathematics for data analysis problems in computer
vision, data mining, machine learning, multimedia, and video surveillance. He
has authored more than 100 scientific articles at top venues including IEEE
T-PAMI, T-IP, ICDM, and CVPR. He received the best theory/algorithm paper
runner up award in IEEE ICDM'07.
\end{IEEEbiography}

\end{document}